\documentclass{article}

\usepackage[main, final]{neurips_2026}

\usepackage[utf8]{inputenc}
\usepackage[T1]{fontenc}
\usepackage{hyperref}
\usepackage{url}
\usepackage{booktabs}
\usepackage{amsfonts}
\usepackage{nicefrac}
\usepackage{microtype}
\usepackage{xcolor}

\usepackage{algorithm}
\usepackage{graphicx}
\usepackage{multirow}
\usepackage{multicol}
\usepackage{kotex}
\usepackage{booktabs}
\usepackage{algpseudocode}
\usepackage[table]{xcolor}
\usepackage{amsmath}
\usepackage{subcaption}
\usepackage{enumitem}

\title{Codebook-Guided Cross-Modal Knowledge Distillation for Structurally Heterogeneous Features}

\author{%
  Dae Ung Jo\\
  Kyungpook National University\\
  \texttt{daeung.jo@knu.ac.kr}\\
  \And
  Jongin Lim \\
  AX/PI Center, Samsung Electronics \\
  \texttt{jonny.lim@samsung.com} \\
  \AND
  YoungJoon Yoo\textsuperscript{*}\\
  Chung-Ang University, SNUAILAB \\
  \texttt{yjyoo3312@cau.ac.kr} \\
  \And
  Daeho Um\textsuperscript{*} \\
  University of Seoul \\
  \texttt{daehoum@uos.ac.kr} \\
}

\begin{document}

\maketitle

\begin{abstract}

Cross-modal knowledge distillation transfers knowledge from a teacher modality to a student modality. Existing feature-level alignment methods typically assume that teacher and student features reside in structurally alignable representation spaces. However, this assumption does not hold when cross-modal features are structurally heterogeneous and lack clear unit-level correspondence, such as 2D spatial visual grids and 1D temporal audio sequences, thereby limiting the applicability of feature-level alignment. To address this challenge, we propose a cross-modal distillation framework that enables effective knowledge transfer across structurally heterogeneous feature spaces via a vector-quantized codebook. Specifically, teacher features are abstracted into a set of vector-form codes regardless of their original feature structure, and the selected codes serve as concept-level anchors for student learning. Code selection is guided by both task relevance and student compatibility, allowing the student to receive transferable teacher knowledge without requiring direct unit-level feature alignment. Experimental results across diverse cross-modal distillation scenarios demonstrate the effectiveness of the proposed framework on classification and semantic segmentation tasks.
\end{abstract}

\begingroup
\renewcommand\thefootnote{}
\footnotetext{\textsuperscript{*}Corresponding authors.}
\addtocounter{footnote}{-1}
\endgroup

\section{Introduction}

Cross-modal knowledge distillation (CMKD) aims to improve a student model operating on a target modality by transferring knowledge from an auxiliary teacher modality available during training~\cite{gupta2016cross}. 
This setting is practically important when an informative modality can provide rich supervision during training, while the deployed model must rely on a single modality at inference time due to sensing constraints, deployment cost, privacy concerns, or missing data~\cite{hoffman2016learning,vapnik2009new}.

Unlike single-modality distillation~\cite{hinton2015distilling}, CMKD transfers knowledge between models built on different modalities, whose inputs, representations, and inductive biases can differ. 
To address this modality gap~\cite{jounderstanding,xue2023modality}, early approaches~\cite{huo2024c2kd} focused on selectively distilling task-relevant information that is beneficial to the student modality, primarily at the prediction level. 
More recent methods~\cite{jang2023stxd,liu2026distilling} have extended this line of work by incorporating not only prediction-level signals but also feature-level cues, such as structural or modality-consistent representations.

While promising, existing approaches inherently assume structural compatibility between teacher and student representations, which enables alignment via projection or shared embedding spaces. 
However, such an assumption does not hold in general cross-modal settings, where representation structures can differ fundamentally. For example, visual features are organized as spatial grids, whereas audio and text features are organized as sequences.
As illustrated in Fig.~\ref{fig:framework}, the same semantic cue can appear as a spatial pattern in one modality and as a sequential pattern in another. 
Directly matching feature units across such heterogeneous structures can impose misleading correspondences. 

In this paper, we address cross-modal knowledge distillation between intermediate representations with distinct structural organizations. 
Our key idea is to avoid direct alignment between raw teacher and student feature units, and instead represent teacher-side intermediate knowledge as a discrete set of transferable anchors. 
Specifically, we learn a vector-quantized teacher codebook over intermediate representations, where the codes serve as anchors that capture recurring task-related patterns in the teacher feature space.
Since not all teacher-side codes are equally useful for the student, we estimate the importance of each code by jointly considering its task relevance and its compatibility with the student modality. 
The selected codes are then used to guide student-side representation learning, allowing the student to exploit fine-grained teacher-side intermediate cues without requiring explicit correspondence between heterogeneous feature units. 

We validate the proposed method across diverse cross-modal distillation settings, including audio-visual, image-text, and RGB-depth scenarios. 
Our experiments cover both classification and segmentation tasks, and compare the proposed framework with prediction-level distillation, feature-level distillation, and modality-gap-aware CMKD baselines. 
The results demonstrate that the proposed framework can effectively leverage intermediate teacher knowledge even when teacher and student features have structurally different organizations.

Our contributions are summarized as follows:
\begin{itemize}[leftmargin=3em]
    \item We address cross-modal knowledge distillation between intermediate representations with distinct structural organizations, where direct feature alignment is unreliable due to the absence of clear unit-level correspondence between teacher and student features.
    
    \item We propose a codebook-guided CMKD framework that learns discrete anchors from teacher intermediate representations, selects transferable teacher-side codes, and uses them to guide student learning without requiring direct feature-unit alignment.
    
    \item We validate the proposed framework across diverse cross-modal benchmarks and tasks, including classification and segmentation, demonstrating its effectiveness over prediction-level, feature-level, and modality-gap-aware distillation baselines.
\end{itemize}
\section{Related Work}

\subsection{Cross-modal Knowledge Distillation}

Knowledge distillation (KD) transfers supervision from a teacher model to a student model. A common form of KD encourages the student to mimic the teacher's predictions by matching output distributions or logits~\cite{hinton2015distilling}. Beyond such prediction-level transfer, many studies have explored richer supervision in the representation space. MLLD~\cite{jin2023multi} improves distillation through multi-level prediction alignment, FitNets~\cite{romero2015fitnetshintsdeepnets} transfers intermediate hint features, RKD~\cite{park2019relational} distills relational structure among samples, CRD~\cite{tian2019contrastive} uses contrastive learning to preserve representation-level knowledge, and OFA~\cite{hao2023one} enables distillation across heterogeneous architectures by projecting intermediate features into an aligned logits-like space. 

More recently, cross-modal knowledge distillation (CMKD) has attracted increasing attention in settings where an auxiliary modality is available during training but not at test time. Compared with conventional KD, CMKD is more challenging because teacher and student are built on different modalities and thus encode both shared semantics and modality-specific factors. In this setting, recent studies have emphasized that effective transfer should account for the modality gap between teacher and student, rather than indiscriminately enforcing the student to imitate all teacher signals~\cite{xue2023modality,jounderstanding}. C2KD explicitly addresses the modality gap through customized distillation strategies for modality imbalance and soft-label misalignment~\cite{huo2024c2kd}. STXD reduces cross-modal discrepancy through structural and temporal distillation~\cite{jang2023stxd}, while FDD disentangles feature components in the frequency domain so that modality-consistent information can be transferred more strongly~\cite{liu2026distilling}. 

Despite these advances, most CMKD methods implicitly assume that teacher and student representations lie in structurally comparable spaces, enabling alignment via projection or shared embeddings.
However, this assumption becomes less straightforward when modalities exhibit substantially different structures.
Consequently, cross-modal distillation across structurally heterogeneous representations remains a challenging problem. In this work, we address this problem with a codebook-based distillation framework that abstracts teacher representations into discrete anchors for student learning.

\subsection{Codebook Learning and Vector Quantization}

Vector quantization (VQ) maps continuous representations to a finite set of learned codewords by assigning each feature to its nearest code. 
This discrete abstraction was popularized by VQ-VAE, which learns a codebook with codebook and commitment objectives~\cite{van2017neural}, and has since been extended through hierarchical quantization and high-resolution visual representation learning~\cite{esser2021taming,razavi2019generating}. 
By replacing continuous features with reusable codewords, VQ provides a compact representation space that can capture recurring patterns in intermediate features. 

Beyond compression and reconstruction, recent studies have explored codebooks as semantic interfaces for multimodal representation learning.
Language-guided codebook learning aligns visual codes with textual semantics~\cite{liang2024lg}, while unified codebook learning has been explored for multimodal large language models~\cite{zheng2024unicode}.
Conceptual codebooks further connect visual representations with conceptual language prompts~\cite{zhang2024conceptual}, and recent work on multimodal unified representations refines discrete code spaces for cross-modal generalization~\cite{huang2025enhancing}.
These studies suggest that learned codebooks can organize continuous feature-level information into compact and semantically meaningful discrete units.

However, these studies do not specifically address cross-modal knowledge distillation between structurally heterogeneous intermediate features, where teacher and student feature units lack direct correspondence.
In contrast, our work uses a teacher-side VQ codebook as a distillation interface, abstracting teacher representations into discrete concept anchors and selecting transferable codes based on task relevance and student compatibility.
This enables feature-level teacher knowledge to guide the student without imposing explicit unit-to-unit alignment across heterogeneous modalities.

\begin{figure*}[t]
    \centering
    \includegraphics[width=\linewidth]{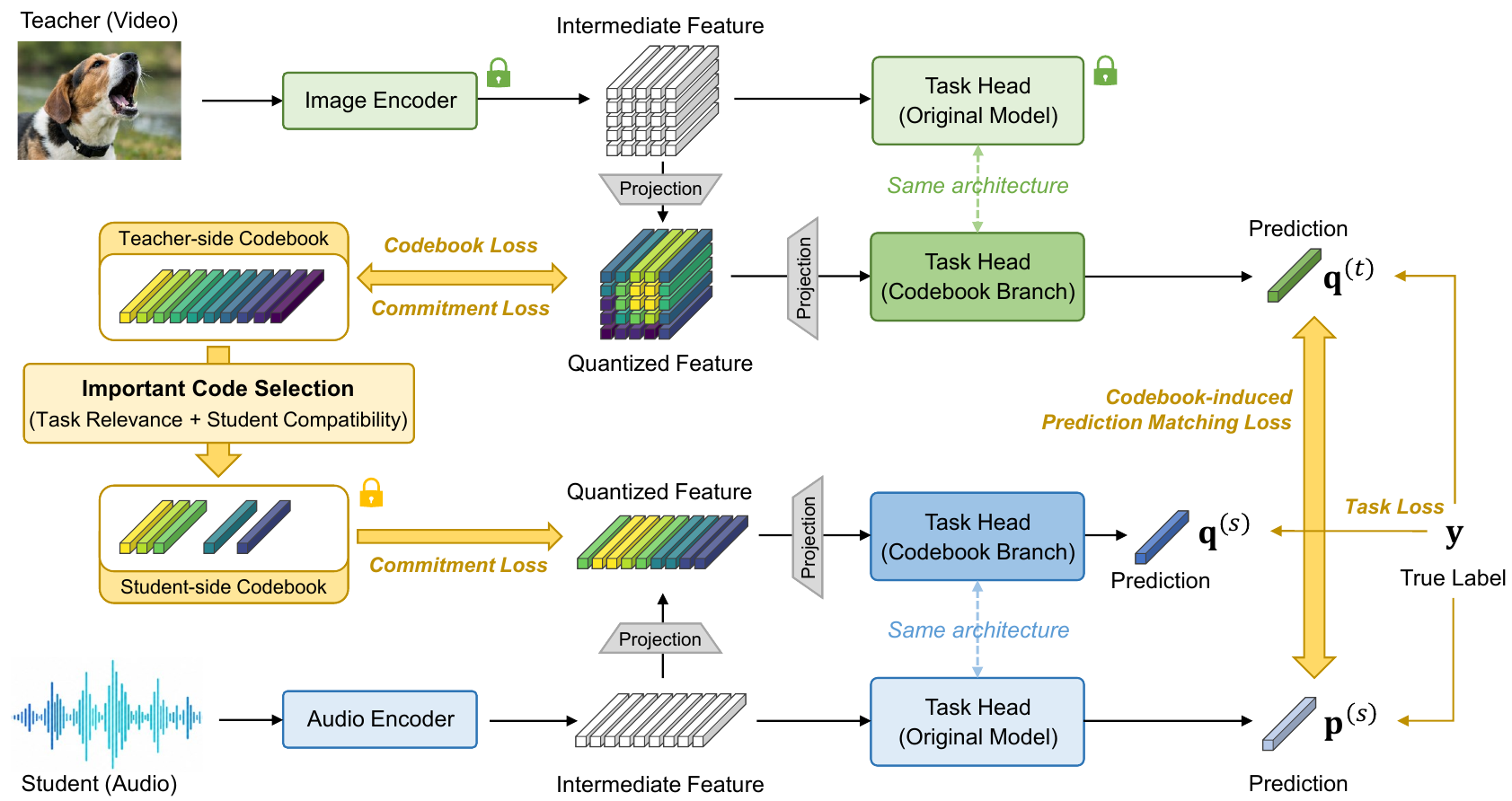}
    \caption{
    Overview of the proposed framework. 
    We learn a teacher-side vector-quantized codebook (Sec.~\ref{subsec:phase1}), select task-relevant and student-compatible codes (Sec.~\ref{subsec:phase2}), and use the selected codes for student-side distillation (Sec.~\ref{subsec:phase3}). 
    The framework enables feature-level CMKD without requiring explicit unit-level correspondence between structurally different teacher and student representations.
    }
    \label{fig:framework}
\end{figure*}

\section{Proposed Method}
\label{sec:method}
\subsection{Notations and Problem Statement}

We consider a cross-modal knowledge distillation setting with paired teacher-modality ($t$) and student-modality ($s$) inputs and corresponding targets,
\begin{equation}
\mathbf{X}^{(t)}=\{\mathbf{x}_i^{(t)}\}_{i=1}^{N}, \qquad
\mathbf{X}^{(s)}=\{\mathbf{x}_i^{(s)}\}_{i=1}^{N}, \qquad
\mathbf{Y}=\{\mathbf{y}_i\}_{i=1}^{N},
\end{equation}
where $N$ denotes the number of training samples. The same subscript $i$ indicates that $\mathbf{x}_i^{(t)}$, $\mathbf{x}_i^{(s)}$, and $\mathbf{y}_i$ correspond to the same sample (i.e., the $i$-th sample). For each modality $m \in \{t,s\}$, let
\begin{equation}
\mathbf{z}_i^{(m)} = f^{(m)}(\mathbf{x}_i^{(m)};\theta^{(m)}) \in \mathbb{R}^{L^{(m)} \times D^{(m)}}, \qquad
\mathbf{p}_i^{(m)} = g^{(m)}(\mathbf{z}_i^{(m)};\phi^{(m)}),
\end{equation}
where $f^{(m)}$ and $g^{(m)}$ denote the modality-specific encoder and task head, respectively. Here, $\mathbf{z}_i^{(m)}$ represents the modality-specific feature representation, and $\mathbf{p}_i^{(m)}$ denotes the corresponding task prediction (e.g., class probabilities in classification). $D^{(m)}$ is the feature dimension and $L^{(m)}$ is the number of modality-specific structured feature units. Because the teacher and student are built on different modalities, their intermediate features may have different structural organizations and sizes. For example, in a 2D visual feature map of image data, $L^{(m)}$ may denote the number of spatial locations (i.e., width $\times$ height), whereas in a 1D sequential feature of audio data, $L^{(m)}$ may denote the number of temporal positions.

Given paired cross-modal training data and a pretrained teacher model $\{\theta^{*(t)}, \phi^{*(t)}\}$, our goal is to learn the student model  $\{\theta^{(s)}, \phi^{(s)}\}$ by transferring useful knowledge from the teacher modality to improve task performance of the student model.

\subsection{Overall Framework}

The proposed CMKD framework for heterogeneously structured features is illustrated in Fig.~\ref{fig:framework}. The proposed framework consists of three stages: teacher-side codebook learning, code-wise importance estimation, and selective codebook-based distillation.
First, we introduce a teacher-side codebook branch and learn a vector-quantized codebook that captures task-related patterns in the teacher feature space. (Sec.~\ref{subsec:phase1})
Second, we estimate the importance of each code by jointly considering task relevance and student compatibility. (Sec.~\ref{subsec:phase2})
Third, we select high-importance codes and use them as a codebook for student-side quantization, enabling feature-level distillation without explicit unit-level correspondence between teacher and student representations. 
We further perform prediction-level alignment between the student and the teacher-side codebook-branch prediction during student training. (Sec.~\ref{subsec:phase3})

\subsection{Stage-1: Teacher-Side Codebook Learning}
\label{subsec:phase1}

In this stage, we freeze the pretrained teacher model with parameters \(\{\theta^{*(t)}, \phi^{*(t)}\}\). 
For each teacher-modality input \(\mathbf{x}_i^{(t)}\), we extract the intermediate feature \(\mathbf{z}_i^{(t)}\) using the fixed teacher encoder:
\begin{equation}
\mathbf{z}_i^{(t)} = f^{(t)}(\mathbf{x}_i^{(t)};\theta^{*(t)}).
\end{equation}
To obtain a discrete abstraction of $\mathbf{z}_i^{(t)}$, we introduce a learnable codebook $\mathbf{C} \in \mathbb{R}^{B \times D}$ with two projection layers, $\Pi_{\mathrm{in}}^{(t)}$ and $\Pi_{\mathrm{out}}^{(t)}$, which project the feature into the code space and reconstruct it back to the original feature space, respectively. The reconstructed feature is then fed into a teacher-side codebook-branch prediction head $g_{\mathrm{cp}}^{(t)}(\cdot;\phi_{\mathrm{cp}}^{(t)})$, which follows the architecture of the original teacher task head but has separate learnable parameters. The overall forward process is given by
\begin{equation}
\mathbf{e}_i^{(t)} = \mathrm{VQ}\Big(\Pi_{\mathrm{in}}^{(t)}(\mathbf{z}_i^{(t)});\mathbf{C}\Big), \qquad
\mathbf{q}_i^{(t)} = g_{\mathrm{cp}}^{(t)}\Big(\Pi_{\mathrm{out}}^{(t)}(\mathbf{e}_i^{(t)});\phi_{\mathrm{cp}}^{(t)}\Big),
\end{equation}
where vector quantization (VQ) assigns each projected token to its nearest codeword in $\mathbf{C}$, thereby converting the continuous teacher feature into a discrete code-based representation~\cite{van2017neural}. Here, $\mathbf{e}_i^{(t)}$ denotes the quantized representation, and $\mathbf{q}_i^{(t)}$ denotes the prediction from the codebook branch.

To train the teacher-side codebook branch, we optimize the task loss together with the standard vector-quantization objective, consisting of the codebook loss and the commitment loss~\cite{van2017neural}:
\begin{equation}
\mathcal{L}_{\mathrm{S1}}^{\mathrm{code}}
= \sum_{i=1}^{N}\bigg(
\mathcal{L}_{\mathrm{task}}(\mathbf{q}_i^{(t)}, \mathbf{y}_i)
+
\left\lVert \operatorname{sg}\!\left[\Pi_{\mathrm{in}}^{(t)}(\mathbf{z}_i^{(t)})\right] - \mathbf{e}_i^{(t)} \right\rVert_2^2
+ \beta_{\mathrm{com}}
\left\lVert \Pi_{\mathrm{in}}^{(t)}(\mathbf{z}_i^{(t)}) - \operatorname{sg}\!\left[\mathbf{e}_i^{(t)}\right] \right\rVert_2^2\bigg),
\end{equation}
where $\operatorname{sg}[\cdot]$ denotes the stop-gradient operator, $\beta_{\mathrm{com}}$ is the commitment loss weight and $\mathcal{L}_{\mathrm{task}}$ denotes the task-specific loss (e.g., cross-entropy for classification).

For effective student distillation, the learned teacher codebook should preserve task-relevant information while remaining compatible with the student modality.
We therefore jointly train the teacher-side codebook branch together with the student model.
Specifically, let $\mathbf{p}_i^{(s)}$ denote the prediction from the original student task head. 
We introduce a \textit{codebook-induced prediction matching} term that aligns $\mathbf{q}_i^{(t)}$ with $\mathbf{p}_i^{(s)}$. 
The overall objective is given by
\begin{equation}
\mathcal{L}_{\mathrm{S1}}
=
\mathcal{L}_{\mathrm{S1}}^{\mathrm{code}}
+ \sum_{i=1}^{N}\bigg(
\mathcal{L}_{\mathrm{task}}(\mathbf{p}_i^{(s)}, \mathbf{y}_i)
+
\left\lVert \mathbf{q}_i^{(t)} - \mathbf{p}_i^{(s)} \right\rVert_2^2\bigg).
\label{eq:phase1}
\end{equation}

\subsection{Stage-2: Code-wise Importance Estimation
}
\label{subsec:phase2}

Although Eq.~\ref{eq:phase1} encourages the teacher-side codebook to learn codes that are compatible with the student modality, not all codes are equally beneficial for transfer. 
We therefore estimate the importance of each code and select a subset of high-importance codes for subsequent distillation.

To estimate code importance, we use the teacher-side codebook branch learned in Stage-1 and evaluate how its prediction changes when each code is removed. 
For each code $k$, we form a \textit{leave-one-code-out} codebook $\mathbf{C}_{-k}$ by removing it from $\mathbf{C}$ and compute the corresponding teacher-side codebook-branch prediction:
\begin{equation}
\bar{\mathbf{e}}_{i}^{(t)} = \mathrm{VQ}\Big(\Pi_{\mathrm{in}}^{(t)}(\mathbf{z}_i^{(t)});\mathbf{C}_{-k}\Big), \qquad
\bar{\mathbf{q}}_{i}^{(t)} = g_{\mathrm{cp}}^{(t)}\Big(\Pi_{\mathrm{out}}^{(t)}(\bar{\mathbf{e}}_{i}^{(t)});\phi_{\mathrm{cp}}^{(t)}\Big).
\end{equation}
We then estimate the importance of the excluded code from two perspectives: task relevance and student compatibility,
\begin{equation}
u_k^{\mathrm{rel}} =
\mathbb{E}_{i}\!\left[
\mathcal{L}_{\mathrm{task}}(\bar{\mathbf{q}}_{i}^{(t)}, \mathbf{y}_i)
-
\mathcal{L}_{\mathrm{task}}(\mathbf{q}_{i}^{(t)}, \mathbf{y}_i)
\right],
u_k^{\mathrm{comp}} =
\mathbb{E}_{i}\!\left[
\left\lVert \bar{\mathbf{q}}_{i}^{(t)} - \mathbf{p}_i^{(s)} \right\rVert_2^2
-
\left\lVert \mathbf{q}_{i}^{(t)} - \mathbf{p}_i^{(s)} \right\rVert_2^2
\right],
\label{eq:score_relevance}
\end{equation}
where $\mathbf{q}_{i}^{(t)}$ is the teacher-side codebook branch prediction obtained with the full codebook $\mathbf{C}$.
The relevance score $u_k^{\mathrm{rel}}$ measures how much removing the $k$-th code degrades the teacher-side task prediction. 
A larger value indicates that the code is more important for preserving task-relevant information. 
The compatibility score $u_k^{\mathrm{comp}}$ measures how much removing the code increases the discrepancy between the teacher-side codebook-branch prediction and the student prediction. 
A larger value indicates that the code provides guidance that is more compatible with the student modality.
After normalizing the two scores, we compute the final code importance by
\begin{equation}
u_k = \,{u}_k^{\mathrm{rel}} + \,{u}_k^{\mathrm{comp}}.
\end{equation}
The resulting scores are used to rank teacher codes and select a subset for the next distillation stage.

\subsection{Stage-3: Selective Codebook-Based Distillation}
\label{subsec:phase3}

In the final stage, we select the top-$K$ codes according to the importance scores obtained in Stage-2 and construct the selected teacher codebooks \(\mathbf{C}_{\mathrm{sel}}\), which is used as teacher-derived anchors for student-side quantization.
Specifically, we build a student codebook branch with learnable projection layers \(\Pi_{\mathrm{in}}^{(s)}\) and \(\Pi_{\mathrm{out}}^{(s)}\), along with a student-side codebook-branch prediction head \(g_{\mathrm{cp}}^{(s)}(\cdot;\phi_{\mathrm{cp}}^{(s)})\). 
Through this branch, student features are quantized using the selected teacher codebook, allowing teacher-side intermediate knowledge to guide student learning without direct feature-unit alignment.

Given the student intermediate feature \(\mathbf{z}_i^{(s)}\), the student-side codebook-branch produces
\begin{equation}
\mathbf{e}_i^{(s)} = \mathrm{VQ}\Big(\Pi_{\mathrm{in}}^{(s)}(\mathbf{z}_i^{(s)});\mathbf{C}_{\mathrm{sel}}\Big), \qquad
\mathbf{q}_i^{(s)} = g_{\mathrm{cp}}^{(s)}\Big(\Pi_{\mathrm{out}}^{(s)}(\mathbf{e}_i^{(s)});\phi_{\mathrm{cp}}^{(s)}\Big),
\end{equation}
where \(\mathbf{C}_{\mathrm{sel}}\) is fixed during this stage. Since the codebook is no longer updated, the codebook branch is trained only with the task loss and the commitment loss:
\begin{equation}
\mathcal{L}_{\mathrm{S3}}^{\mathrm{code}}
=
\mathcal{L}_{\mathrm{task}}(\mathbf{q}_i^{(s)}, \mathbf{y}_i)
+ \beta_{\mathrm{com}}
\left\lVert
\Pi_{\mathrm{in}}^{(s)}(\mathbf{z}_i^{(s)})
-
\operatorname{sg}\!\left[\mathbf{e}_i^{(s)}\right]
\right\rVert_2^2.
\label{eq:phase3_codebook}
\end{equation}
The teacher-side codebook branch learned in Stage-1 produces a codebook-induced prediction that reflects quantized teacher knowledge.
To fully leverage the learned teacher codebook, we use its induced prediction to guide the student task prediction, allowing the codebook to jointly shape student-side quantization and final prediction learning.
Specifically, we introduce a \textit{codebook-induced prediction matching} term between \(\mathbf{q}_{i}^{(t)}\) and \(\mathbf{p}_i^{(s)}\), leading to the Stage-3 objective:
\begin{equation}
\mathcal{L}_{\mathrm{S3}}
=
\mathcal{L}_{\mathrm{task}}(\mathbf{p}_i^{(s)}, \mathbf{y}_i)
+
\lambda_{\mathrm{dist}}
\bigg(
\mathcal{L}_{\mathrm{S3}}^{\mathrm{code}}
+
\left\lVert \mathbf{q}_{i}^{(t)} - \mathbf{p}_i^{(s)} \right\rVert_2^2
\bigg),
\label{eq:phase3}
\end{equation}
where $\lambda_{\mathrm{dist}}$ controls the contribution of the distillation terms. 
Through this stage, the student is trained with the original task objective and two codebook-based distillation signals, which guide both student intermediate representations and task predictions using the learned teacher codebook, without requiring direct alignment between structurally heterogeneous features. 
Algorithm~\ref{alg:proposed} in Appendix~\ref{appendix:algorithm} provides the pseudo-code of the proposed method.

\begin{table*}[!t]
\small
\caption{
Classification results. 
A, V, T, and I denote audio, video, text, and image modalities, respectively, and each arrow indicates the teacher-to-student distillation direction. 
We report accuracy averaged over five independent runs, with the best result in each column shown in bold.
}
\label{tab:cls_result}
\centering
\begin{tabular}{l|cccccccc}
\toprule
\multirow{2}{*}{Method}& \multicolumn{2}{c}{RAVDESS} & \multicolumn{2}{c}{VGG-Sound} & \multicolumn{2}{c}{CrisisMMD} & \multicolumn{2}{c}{MM-IMDB}\\
 & A $\rightarrow$ V & V $\rightarrow$ A & A $\rightarrow$ V & V $\rightarrow$ A & T $\rightarrow$ I & I $\rightarrow$ T & T $\rightarrow$ I & I $\rightarrow$ T\\
\midrule
w/o KD &0.9208	&0.6826 &0.5031&0.6058 &0.5403&0.5500 &0.6645&0.7294\\
\midrule
\multicolumn{9}{c}{\textit{General distillation baselines}} \\
\midrule
KLD     &0.9118&0.6972  &0.5153&0.6176 &0.5377&0.5578 &0.6732&0.7310\\
MLLD    &0.9292&0.6889  &0.5135&0.6122 &0.5357&0.5562 &0.6699&0.7320\\
FitNets &0.9062&0.7146  &0.4916&0.5962 &0.5281&0.5507 &0.6759&0.7296\\
RKD     &0.9236&0.7014  &0.5066&0.6085 &0.5432&0.5569 &0.6668&0.7317\\
CRD     &0.9299&0.7007  &0.5063&0.6110 &0.5377&0.5560 &0.6685&0.7299\\
OFA     &0.9215&0.6562  &0.5100&0.6129 &0.5362&0.5565 &0.6715&0.7339\\
\midrule
\multicolumn{9}{c}{\textit{Modality-gap-aware CMKD baselines}} \\
\midrule
MGDFR   &0.9313&0.7264  &0.5225&0.6186 &\textbf{0.5440}&0.5584 &0.6725&0.7382\\
C2KD    &0.9306&0.7021  &0.5123&0.6154 &0.5410&0.5520 &0.6670&0.7372\\
STXD    &0.9035&0.7250  &0.5067&0.6093 &0.5421&0.5485 &0.6761&0.7265\\
FDD     &0.9153&0.7055  &0.5028&0.6063 &0.5382&0.5547 &0.6670&0.7323\\
\midrule
Ours    &\textbf{0.9431}&\textbf{0.7486}  &\textbf{0.5299}&\textbf{0.6229} &0.5433&\textbf{0.5592} &\textbf{0.6768}&\textbf{0.7434}\\ 
\bottomrule
\end{tabular}
\end{table*}

\section{Experiments}
We validate the proposed method on five multimodal benchmarks spanning classification and semantic segmentation tasks in CMKD settings. Implementation  details are provided in the Appendix~\ref{app:implementation_details}.

\subsection{Classification Experiments}
\label{exp:classification}
\paragraph{Datasets.}
We use four multimodal classification benchmarks. 
RAVDESS~\cite{livingstone2018ryerson} contains paired speech audio and facial video recordings for eight-class emotion recognition. 
VGG-Sound~\cite{chen2020vggsound} is a video--audio event classification dataset, where we use the curated 141-class subset following \cite{li2025mst}. 
CrisisMMD-v2.0~\cite{alam2018crisismmd,ofli2020analysis} consists of disaster-related image--text social media posts for eight-class humanitarian category classification. 
MM-IMDB~\cite{arevalo2017gated} contains paired movie posters and plot descriptions for movie genre classification. 
For all datasets, we use a 6:2:2 train/validation/test split and report test accuracy from the best validation checkpoint, averaged over five runs.
\paragraph{Baselines.} 
We compare our method with general distillation methods including KLD~\cite{hinton2015distilling} and MLLD~\cite{jin2023multi}, FitNets~\cite{romero2015fitnetshintsdeepnets}, RKD~\cite{park2019relational}, CRD~\cite{tian2019contrastive}, and OFA~\cite{hao2023one}; and modality-gap-aware CMKD methods, including MGDFR~\cite{xue2023modality}, C2KD~\cite{huo2024c2kd}, STXD~\cite{jang2023stxd}, and FDD~\cite{liu2026distilling}.
For feature-alignment baselines, which are not directly applicable to structurally heterogeneous features, we use pooled representations and add projection layers when feature dimensions differ.
\paragraph{Implementation.}
Image and video inputs are represented as 2D spatial feature  using ResNet- and ViT-based feature extractors~\cite{dosovitskiy2020image,he2016deep}. 
Audio inputs are represented as 1D temporal feature sequences using Wav2Vec2- and VGGish-style feature extractors~\cite{baevski2020wav2vec,hershey2017cnn}. 
Text inputs are represented as token-level sequence features using transformer-based feature extractors including BERTweet and FLAVA~\cite{nguyen2020bertweet,singh2022flava}. 
We select the top 50\% of teacher codes according to the proposed importance score.
\paragraph{Results.}
Tab.~\ref{tab:cls_result} summarizes the classification results.
Our method achieves the best performance in seven out of eight settings and the second-best result on CrisisMMD T $\rightarrow$ I.
It also improves over the student trained without distillation in all transfer directions, showing that the proposed framework provides useful supervision across both audio-visual and text-image scenarios.
Modality-gap-aware CMKD baselines are generally competitive, which supports the importance of accounting for cross-modal discrepancy.
Nevertheless, our method obtains stronger results in most settings, suggesting that useful teacher-side cues can be extracted from structurally heterogeneous intermediate features and transferred through a codebook-level interface.

\begin{table}[t]
\centering
\caption{Semantic segmentation results on NYU-Depth. 
\textbf{Left}: quantitative comparison for two distillation scenarios, where the first modality denotes the teacher and the second denotes the student. 
\textbf{Right}: class-wise recall comparison of RGB-only, depth-only, and our RGB $\rightarrow$ Depth student models.}
\vspace{1.5mm}
\begin{minipage}[t]{0.67\linewidth}
\vspace{0pt}
\centering
\small
\resizebox{\linewidth}{!}{
\begin{tabular}{l|cccccc}
\toprule
\multirow{2}{*}{Method} & \multicolumn{3}{c}{Depth $\rightarrow$ RGB} & \multicolumn{3}{c}{RGB $\rightarrow$ Depth}  \\
& OA $\uparrow$ & CA $\uparrow$ & mIoU $\uparrow$ & OA $\uparrow$ & CA $\uparrow$ & mIoU $\uparrow$\\
\midrule
w/o KD  &0.5582&0.2748&0.1850&\textbf{0.5269}&0.2100&0.1366 \\
\midrule
MLLD    &0.5599&0.2792&0.1911&0.5250&0.2118&0.1399\\
CRD     &0.5592&0.2677&0.1806&0.5261&0.2111&0.1385\\
OFA     &0.5607&0.2881&0.1982&0.5242&0.2123&0.1409\\
MGDFR   &0.5601&0.2737&0.1878&0.5213&0.2162&0.1442\\
C2KD	&0.5587&0.2771&0.1878&0.5214&0.2093&0.1396\\
STXD    &0.5616&0.2786&0.1925&0.5231&0.2164&0.1425\\
FDD     &0.5608&0.2744&0.1855&0.5252&0.2122&0.1403\\
\midrule
Ours     &\textbf{0.5632}&\textbf{0.2909}&\textbf{0.2003}&0.5218&\textbf{0.2184}&\textbf{0.1456}\\
\bottomrule
\end{tabular}
}
\end{minipage}
\hfill
\begin{minipage}[t]{0.32\linewidth}
\vspace{0pt}
\centering
\includegraphics[width=\linewidth]{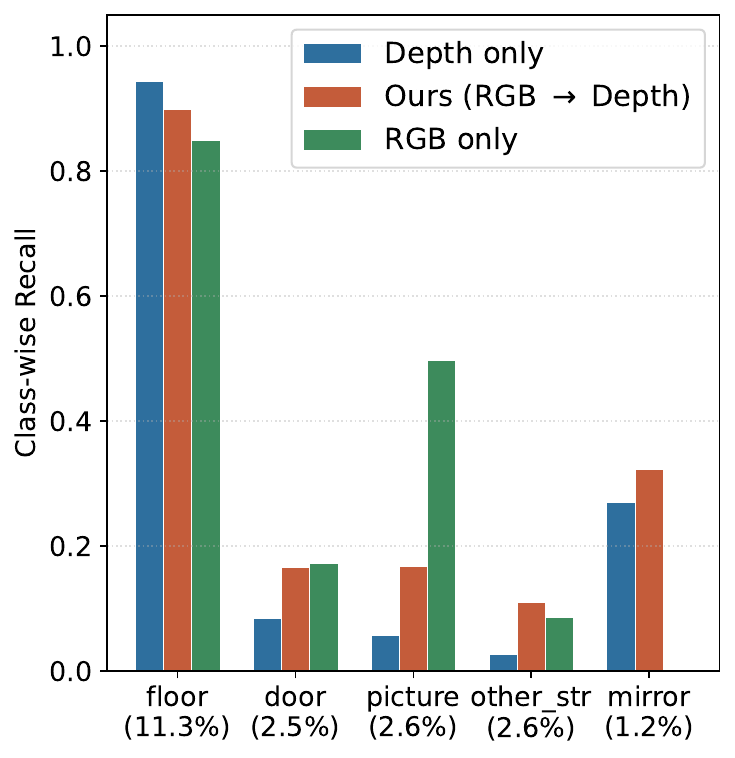}
\end{minipage}
\label{tab:seg_result}
\end{table}

\subsection{Semantic Segmentation Experiments}

\paragraph{Dataset.}
We evaluate semantic segmentation on NYU-Depth V2~\cite{silberman2012indoor}, an indoor RGB-D benchmark with aligned RGB-depth image pairs and pixel-wise annotations for 40 classes.
Following~\cite{li2025mst}, we report Overall Accuracy (OA), Class-averaged Accuracy (CA), and mean Intersection-over-Union (mIoU), which measure pixel-level accuracy, average class-wise recall, and average region overlap, respectively.
We use a 6:2:2 train/validation/test split and report results averaged over five runs.

\paragraph{Baselines.} 
We use representative baselines from the classification experiments.
Since RGB and depth inputs are spatially aligned, dense correspondence between teacher and student feature maps is available.
Thus, feature-level baselines are applied directly to corresponding intermediate feature maps when applicable, with projection layers added only to match channel dimensions.

\paragraph{Implementation.}
We use a ResNet-34~\cite{he2016deep} based feature extractor for both RGB and depth modalities, and apply our method to the resulting $14 \times 14$ grid features. 
The codebook contains 1024 codes, from which the top 12.5\% are selected based on the estimated importance scores. 

\paragraph{Results.}
Tab.~\ref{tab:seg_result} reports the semantic segmentation results.
The left part provides quantitative comparisons, and the right part shows class-wise recall for selected categories in the RGB$\rightarrow$Depth setting.
Our method achieves the best performance on most metrics, with the only exception being OA in RGB$\rightarrow$Depth, where the w/o KD baseline remains slightly higher.
Since OA is weighted by pixel frequency, recall drops in high-occupancy categories can noticeably affect the overall score (Appendix~\ref{app:nyu_classwise}).
The decrease on \texttt{floor} in the class-wise recall comparison provides one representative example.
Meanwhile, our method improves CA and mIoU, with recall gains on categories such as \texttt{door}, \texttt{picture}, and \texttt{other\_structure}, where depth-only features may provide less discriminative cues but can benefit from guidance provided by the RGB teacher.
These results suggest that the proposed method can transfer useful teacher-side information for class-level segmentation quality.

\begin{figure*}[t]
    \centering
    \includegraphics[width=\linewidth]{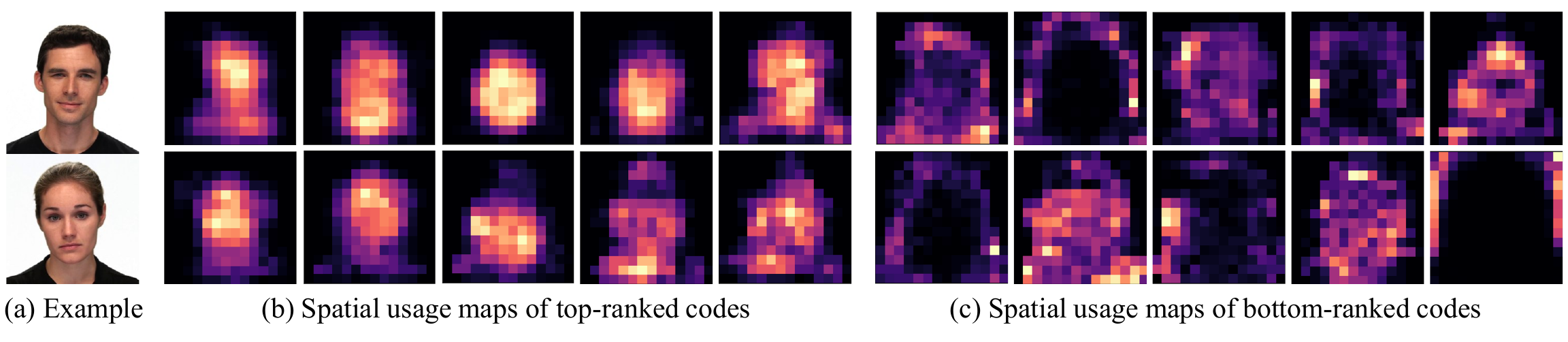}
        \caption{
        Codebook analysis on RAVDESS.
        (a) Example frames from the dataset.
        (b,c) Spatial usage maps of top-ranked and bottom-ranked codes according to the proposed importance score.
        }
    \label{fig:codebook_analysis}
    \vspace{-3mm}
\end{figure*}

\vspace{2mm}
\begin{table}[th!]
\small
\centering
\caption{Ablation study on Stage-3 loss components.
``Code'' denotes the student-side codebook-branch objective.
``Pred-Match'' denotes the codebook-induced prediction matching term.
}
\vspace{1mm}
\label{tab:loss_ablation}
\begin{tabular}{cc|cccccccc}
\toprule
\multirow{2}{*}{Code} & Pred & \multicolumn{2}{c}{RAVDESS} & \multicolumn{2}{c}{VGG-Sound} & \multicolumn{2}{c}{CrisisMMD} & \multicolumn{2}{c}{MM-IMDB}\\
&Match& A $\rightarrow$ V & V $\rightarrow$ A & A $\rightarrow$ V & V $\rightarrow$ A & T $\rightarrow$ I & I $\rightarrow$ T & T $\rightarrow$ I & I $\rightarrow$ T\\
\midrule
-&- 
&0.9208&0.6826 &0.5031&0.6058 &0.5403&0.5500 &0.6645&0.7294\\
\checkmark &-
&0.9202&0.7333 &0.5067&0.6098  &0.5387&0.5534 &0.6716&0.7363\\
-& \checkmark            
&0.9285&0.7479 &0.5294&0.6204 &0.5367&0.5529 &0.6770&0.7421\\
\checkmark & \checkmark 
&0.9431&0.7486 &0.5299&0.6229 &0.5433&0.5592 &0.6768&0.7434\\
\bottomrule
\end{tabular}
\end{table}

\begin{table}[th!]
\vspace{-4mm}
\small
\centering
\caption{Ablation on the prediction matching target in Stage-3 Loss. 
}
\vspace{1mm}
\label{tab:pred_matching_target}
\resizebox{\linewidth}{!}{
\begin{tabular}{c|cccccccc}
\toprule
\multirow{2}{*}{Pred-Match Target} & \multicolumn{2}{c}{RAVDESS} & \multicolumn{2}{c}{VGG-Sound} & \multicolumn{2}{c}{CrisisMMD} & \multicolumn{2}{c}{MM-IMDB}\\
& A $\rightarrow$ V & V $\rightarrow$ A & A $\rightarrow$ V & V $\rightarrow$ A & T $\rightarrow$ I & I $\rightarrow$ T & T $\rightarrow$ I & I $\rightarrow$ T\\
\midrule
Original branch $(\mathbf{p}^{(t)})$
&0.8903&0.6201 &0.5173&0.6171 &0.5425&0.5539 &0.6654&0.7351\\
Codebook branch $(\mathbf{q}^{(t)})$
&0.9431&0.7486 &0.5299&0.6229 &0.5433&0.5592 &0.6768&0.7434\\
\bottomrule
\end{tabular}
}
\end{table}

\subsection{Analysis}

\paragraph{Do High-Importance Codes Capture Informative Patterns?}
We further analyze the learned codebook on RAVDESS, where the actor faces are well aligned across videos, making the spatial usage patterns of codes easier to interpret.
We consider the visual-teacher setting and visualize where each code is assigned over the spatial feature grid across the dataset.
Fig.~\ref{fig:codebook_analysis} shows the usage maps of the top-10 and bottom-10 codes according to the importance score.
Since RAVDESS requires emotion recognition from acted facial and vocal expressions, task-relevant visual concepts are expected to be concentrated around centered regions, especially the face and mouth.
Consistent with this intuition, high-score codes are mainly activated around the central regions, suggesting that they capture visual concepts that are informative to the task and likely useful for cross-modal transfer.
In contrast, low-score codes tend to show more scattered or peripheral usage patterns, indicating that they are less consistently associated with meaningful regions.

\paragraph{Effect of Distillation Loss Components.}
Tab.~\ref{tab:loss_ablation} analyzes the contribution of the two main Stage-3 loss components in Eq.~\ref{eq:phase3}.
``Code'' denotes the student-side codebook-branch objective in Eq.~\eqref{eq:phase3_codebook}, which constrains student features through the selected teacher codebook and provides representation-level guidance.
``Pred-Match'' denotes the codebook-induced prediction matching term in Eq.~\eqref{eq:phase3}, which guides the student task prediction using the teacher prediction derived from quantized representations.
While each component improves several transfer directions, combining them yields the strongest overall performance, suggesting complementary effects between representation guidance and codebook-induced prediction guidance.

Tab.~\ref{tab:pred_matching_target} further examines the target used for prediction matching by comparing the original teacher prediction $\mathbf{p}^{(t)}$ with the codebook-induced teacher prediction $\mathbf{q}^{(t)}$.
Matching $\mathbf{q}^{(t)}$ outperforms matching $\mathbf{p}^{(t)}$ across all evaluated datasets and transfer directions, indicating that the gain from ``Pred-Match'' does not come from generic output-level supervision alone, but from using a guidance signal shaped by the learned teacher codebook.

\begin{figure*}[t]
    \centering
    \vspace{-2mm}
    \includegraphics[width=\linewidth]{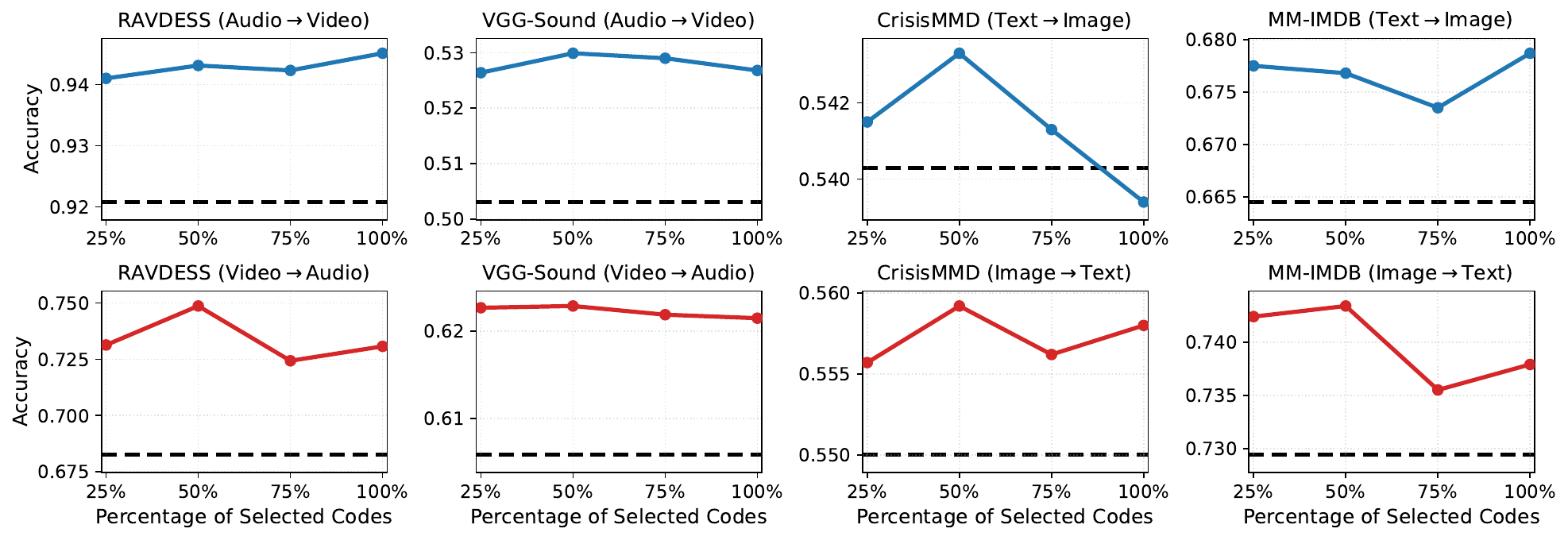}
    \caption{
    Selected code ratio ablation.
    We vary the percentage of selected teacher codes used for student-side quantization from 25\% to 100\%.
    The dashed line denotes the w/o KD baseline.
    }
    \vspace{-4mm}
    \label{fig:selected_code_ablation}
\end{figure*}

\paragraph{Effect of Code Selection.}
Fig.~\ref{fig:selected_code_ablation} analyzes the effect of the selected code ratio by varying the percentage of teacher codes used for student-side quantization from 25\% to 100\%, where the dashed line denotes the w/o KD baseline.
The proposed method remains above the w/o KD baseline in most cases, indicating that its benefit is not limited to a single carefully tuned ratio.
The best ratio varies across settings, suggesting that the amount of useful teacher-side code information can depend on the dataset and transfer direction.
In the main classification experiments in Sec.~\ref{exp:classification}, we use a fixed ratio of 50\% for all settings to maintain a simple and consistent evaluation protocol (Appendix~\ref{app:retention}).

Tab.~\ref{tab:score_ablation} further analyzes the effect of different code selection criteria under the fixed 50\% selection ratio used in Sec.~\ref{exp:classification}.
Compared with random selection, score-based criteria generally improve performance, indicating that the learned scores provide useful signals for selecting teacher codes.
When codes are selected based only on task relevance or only on student compatibility score, the two criteria show different strengths across scenarios.
Using their combined score achieves the best or competitive performance in most cases, suggesting that jointly considering task relevance and student compatibility provides a more balanced criterion for identifying useful teacher codes.

\begin{table}[t]
\small
\centering
\caption{Ablation study on code selection criteria. 
}
\vspace{1mm}
\label{tab:score_ablation}
\resizebox{\linewidth}{!}{
\begin{tabular}{l|cccccccc}
\toprule
\multirow{2}{*}{Selection Criteria} & \multicolumn{2}{c}{RAVDESS} & \multicolumn{2}{c}{VGG-Sound} & \multicolumn{2}{c}{CrisisMMD} & \multicolumn{2}{c}{MM-IMDB}\\
& A $\rightarrow$ V & V $\rightarrow$ A & A $\rightarrow$ V & V $\rightarrow$ A & T $\rightarrow$ I & I $\rightarrow$ T & T $\rightarrow$ I & I $\rightarrow$ T\\
\midrule
Random
&0.9292&0.7396 &0.5252&0.6211 &0.5353&0.5536 &0.6718&0.7353\\
Task relevance only
&0.9396&0.7361 &0.5276&0.6233 &0.5435&0.5558 &0.6727&0.7351\\
Student compatibility only
&0.9375&0.7417 &0.5275&0.6222 &0.5437&0.5565 &0.6722&0.7346\\
Combination (Ours)
&0.9431&0.7486 &0.5299&0.6229 &0.5433&0.5592 &0.6768&0.7434\\
\bottomrule
\end{tabular}
}
\end{table}

\section{Conclusion}

We presented a codebook-guided framework for cross-modal knowledge distillation between structurally heterogeneous features.
Instead of directly aligning teacher and student feature units, our method abstracts teacher representations into a vector-quantized codebook and selects task-relevant and student-compatible codes for student learning.
The selected codes are then fully leveraged through student-side codebook quantization and codebook-induced prediction matching.
This design allows the student to exploit fine-grained teacher-side information without requiring explicit unit-level correspondence between heterogeneous feature structures, such as spatial grids and temporal sequences.
Across diverse cross-modal classification and semantic segmentation settings, our method achieves strong performance compared with general and modality-gap-aware distillation baselines.
Further analyses support the proposed framework design and suggest that codebook-guided distillation is a practical strategy for transferring knowledge across structurally heterogeneous features.

\textbf{Future work.} $\quad$In the current framework, code selection relies on a fixed ratio and an offline importance estimation stage.
The fixed ratio may not capture the varying amount of useful teacher-side information across cross-modal scenarios, while offline scoring can increase the cost for larger codebooks (Appendix~\ref{app:comp_cost}).
Future work will explore adaptive and efficient strategies for selecting appropriate code subsets according to the distillation scenario.

\section*{Acknowledgments}
This research was supported by Kyungpook National University Research Fund, 2025.
This work was supported by the National Research Foundation of Korea (NRF) grant funded by the Korea government (MSIT) (RS-2026-25495369).
This work was supported by the Institute of Information \& Communications Technology Planning \& Evaluation (IITP) grant funded by the Korea government (MSIT) [RS-2021-II211341, Artificial Intelligence Graduate School Program (Chung-Ang University), and RS-2022-II220124, Development of Artificial Intelligence Technology for Self-Improving Competency-Aware Learning Capabilities].
This research was also supported by the AI Seoul Tech Research Support Program of the Seoul Future Foundation.

{
\small
\bibliographystyle{plain}
\bibliography{ref.bib}

@article{hinton2015distilling,
  title={Distilling the knowledge in a neural network},
  author={Hinton, Geoffrey and Vinyals, Oriol and Dean, Jeff},
  journal={arXiv preprint arXiv:1503.02531},
  year={2015}
}

@article{liang2024lg,
  title={Lg-vq: Language-guided codebook learning},
  author={Liang, Guotao and Zhang, Baoquan and Wang, Yaowei and Li, Xutao and Ye, Yunming and Wang, Huaibin and Luo, Chuyao and Ye, Kola and others},
  journal={arXiv preprint arXiv:2405.14206},
  year={2024}
}

@inproceedings{zheng2024unicode,
  title={Unicode: Learning a unified codebook for multimodal large language models},
  author={Zheng, Sipeng and Zhou, Bohan and Feng, Yicheng and Wang, Ye and Lu, Zongqing},
  booktitle={European Conference on Computer Vision},
  pages={426--443},
  year={2024},
  organization={Springer}
}

@inproceedings{zhang2024conceptual,
  title={Conceptual codebook learning for vision-language models},
  author={Zhang, Yi and Yu, Ke and Wu, Siqi and He, Zhihai},
  booktitle={European Conference on Computer Vision},
  pages={235--251},
  year={2024},
  organization={Springer}
}

@inproceedings{huang2025enhancing,
  title={Enhancing multimodal unified representations for cross modal generalization},
  author={Huang, Hai and Xia, Yan and Ji, Shengpeng and Wang, Shulei and Wang, Hanting and Fang, Minghui and Zhu, Jieming and Dong, Zhenhua and Zhou, Sashuai and Zhao, Zhou},
  booktitle={Findings of the Association for Computational Linguistics: ACL 2025},
  pages={2353--2366},
  year={2025}
}

@article{van2017neural,
  title={Neural discrete representation learning},
  author={Van Den Oord, Aaron and Vinyals, Oriol and others},
  journal={Advances in neural information processing systems},
  volume={30},
  year={2017}
}

@article{razavi2019generating,
  title={Generating diverse high-fidelity images with vq-vae-2},
  author={Razavi, Ali and Van den Oord, Aaron and Vinyals, Oriol},
  journal={Advances in neural information processing systems},
  volume={32},
  year={2019}
}

@inproceedings{esser2021taming,
  title={Taming transformers for high-resolution image synthesis},
  author={Esser, Patrick and Rombach, Robin and Ommer, Bjorn},
  booktitle={Proceedings of the IEEE/CVF conference on computer vision and pattern recognition},
  pages={12873--12883},
  year={2021}
}

@article{dosovitskiy2020image,
  title={An image is worth 16x16 words: Transformers for image recognition at scale},
  author={Dosovitskiy, Alexey and Beyer, Lucas and Kolesnikov, Alexander and Weissenborn, Dirk and Zhai, Xiaohua and Unterthiner, Thomas and Dehghani, Mostafa and Minderer, Matthias and Heigold, Georg and Gelly, Sylvain and others},
  journal={arXiv preprint arXiv:2010.11929},
  year={2020}
}

@inproceedings{deng2009imagenet,
  title={Imagenet: A large-scale hierarchical image database},
  author={Deng, Jia and Dong, Wei and Socher, Richard and Li, Li-Jia and Li, Kai and Fei-Fei, Li},
  booktitle={2009 IEEE conference on computer vision and pattern recognition},
  pages={248--255},
  year={2009},
  organization={Ieee}
}

@inproceedings{liu2026distilling,
  title={Distilling Cross-Modal Knowledge via Feature Disentanglement},
  author={Liu, Junhong and Zhang, Yuan and Huang, Tao and Xu, Wenchao and Yang, Renyu},
  booktitle={Proceedings of the AAAI Conference on Artificial Intelligence},
  volume={40},
  number={28},
  pages={23739--23747},
  year={2026}
}

@inproceedings{jin2023multi,
  title     = {Multi-Level Logit Distillation},
  author    = {Jin, Ying and Wang, Jiaqi and Lin, Dahua},
  booktitle = {Proceedings of the IEEE/CVF Conference on Computer Vision and Pattern Recognition (CVPR)},
  pages     = {24276--24285},
  year      = {2023}
}

@inproceedings{park2019relational,
  title     = {Relational Knowledge Distillation},
  author    = {Park, Wonpyo and Kim, Dongju and Lu, Yan and Cho, Minsu},
  booktitle = {Proceedings of the IEEE/CVF Conference on Computer Vision and Pattern Recognition (CVPR)},
  pages     = {3967--3976},
  year      = {2019}
}

@article{hao2023one,
  title   = {One-for-All: Bridge the Gap Between Heterogeneous Architectures in Knowledge Distillation},
  author  = {Hao, Zhiwei and Guo, Jianyuan and Han, Kai and Tang, Yehui and Hu, Han and Wang, Yunhe and Xu, Chang},
  journal = {Advances in Neural Information Processing Systems},
  volume  = {36},
  pages   = {79570--79582},
  year    = {2023}
}

@article{romero2015fitnetshintsdeepnets,
  title   = {FitNets: Hints for Thin Deep Nets},
  author  = {Romero, Adriana and Ballas, Nicolas and Kahou, Samira Ebrahimi and Chassang, Antoine and Gatta, Carlo and Bengio, Yoshua},
  journal = {arXiv preprint arXiv:1412.6550},
  year    = {2015}
}

@article{tian2019contrastive,
  title   = {Contrastive Representation Distillation},
  author  = {Tian, Yonglong and Krishnan, Dilip and Isola, Phillip},
  journal = {arXiv preprint arXiv:1910.10699},
  year    = {2019}
}

@inproceedings{huo2024c2kd,
  title     = {C2KD: Bridging the Modality Gap for Cross-Modal Knowledge Distillation},
  author    = {Huo, Fushuo and Xu, Wenchao and Guo, Jingcai and Wang, Haozhao and Guo, Song},
  booktitle = {Proceedings of the IEEE/CVF Conference on Computer Vision and Pattern Recognition (CVPR)},
  pages     = {16006--16015},
  year      = {2024}
}

@article{jang2023stxd,
  title   = {STXD: Structural and Temporal Cross-Modal Distillation for Multi-View 3D Object Detection},
  author  = {Jang, Sujin and Jo, Dae Ung and Hwang, Sung Ju and Lee, Dongwook and Ji, Daehyun},
  journal = {Advances in Neural Information Processing Systems},
  volume  = {36},
  pages   = {29323--29342},
  year    = {2023}
}

@inproceedings{xue2023modality,
  title     = {The Modality Focusing Hypothesis: Towards Understanding Cross-Modal Knowledge Distillation},
  author    = {Xue, Zihui and Gao, Zhengqi and Ren, Sucheng and Zhao, Hang},
  booktitle = {Proceedings of the International Conference on Learning Representations (ICLR)},
  year      = {2023}
}

@article{jounderstanding,
  title   = {Understanding Dimensional Collapse in Cross-Modal Feature Distillation},
  author  = {Jo, Dae Ung and Jang, Sujin and Heo, Jay and Hwang, Sung Ju},
  journal = {OpenReview},
  note    = {ICLR 2025 submission},
  year    = {2025}
}

@inproceedings{li2025mst,
  title     = {MST-Distill: Mixture of Specialized Teachers for Cross-Modal Knowledge Distillation},
  author    = {Li, Hui and Yang, Pengfei and Chen, Juanyang and Dong, Le and Chen, Yanxin and Wang, Quan},
  booktitle = {Proceedings of the 33rd ACM International Conference on Multimedia},
  pages     = {1588--1597},
  year      = {2025}
}

@article{livingstone2018ryerson,
  title={The Ryerson Audio-Visual Database of Emotional Speech and Song (RAVDESS): A dynamic, multimodal set of facial and vocal expressions in North American English},
  author={Livingstone, Steven R and Russo, Frank A},
  journal={PloS one},
  volume={13},
  number={5},
  pages={e0196391},
  year={2018},
  publisher={Public Library of Science San Francisco, CA USA}
}

@inproceedings{chen2020vggsound,
  title={Vggsound: A large-scale audio-visual dataset},
  author={Chen, Honglie and Xie, Weidi and Vedaldi, Andrea and Zisserman, Andrew},
  booktitle={ICASSP 2020-2020 IEEE International Conference on Acoustics, Speech and Signal Processing (ICASSP)},
  pages={721--725},
  year={2020},
  organization={IEEE}
}

@inproceedings{alam2018crisismmd,
  title={Crisismmd: Multimodal twitter datasets from natural disasters},
  author={Alam, Firoj and Ofli, Ferda and Imran, Muhammad},
  booktitle={Proceedings of the international AAAI conference on web and social media},
  volume={12},
  number={1},
  year={2018}
}

@inproceedings{arevalo2017gated,
  title={Gated Multimodal Units for Information Fusion},
  author={Arevalo, John and Solorio, Thamar and Montes-y-G{\'o}mez, Manuel and Gonz{\'a}lez, Fabio A.},
  booktitle={Proceedings of the International Conference on Learning Representations Workshop},
  year={2017}
}

@article{zhou2022contrastive,
  title={Contrastive positive sample propagation along the audio-visual event line},
  author={Zhou, Jinxing and Guo, Dan and Wang, Meng},
  journal={IEEE Transactions on Pattern Analysis and Machine Intelligence},
  volume={45},
  number={6},
  pages={7239--7257},
  year={2022},
  publisher={IEEE}
}

@inproceedings{silberman2012indoor,
  title={Indoor segmentation and support inference from rgbd images},
  author={Silberman, Nathan and Hoiem, Derek and Kohli, Pushmeet and Fergus, Rob},
  booktitle={European conference on computer vision},
  pages={746--760},
  year={2012},
  organization={Springer}
}

@inproceedings{nguyen2020bertweet,
  title={BERTweet: A pre-trained language model for English Tweets},
  author={Nguyen, Dat Quoc and Vu, Thanh and Nguyen, Anh-Tuan},
  booktitle={Proceedings of the 2020 conference on empirical methods in natural language processing: system demonstrations},
  pages={9--14},
  year={2020}
}

@inproceedings{he2016deep,
  title={Deep residual learning for image recognition},
  author={He, Kaiming and Zhang, Xiangyu and Ren, Shaoqing and Sun, Jian},
  booktitle={Proceedings of the IEEE conference on computer vision and pattern recognition},
  pages={770--778},
  year={2016}
}

@article{simonyan2014very,
  title={Very deep convolutional networks for large-scale image recognition},
  author={Simonyan, Karen and Zisserman, Andrew},
  journal={arXiv preprint arXiv:1409.1556},
  year={2014}
}

@article{baevski2020wav2vec,
  title={wav2vec 2.0: A framework for self-supervised learning of speech representations},
  author={Baevski, Alexei and Zhou, Yuhao and Mohamed, Abdelrahman and Auli, Michael},
  journal={Advances in neural information processing systems},
  volume={33},
  pages={12449--12460},
  year={2020}
}

@inproceedings{hershey2017cnn,
  title={CNN architectures for large-scale audio classification},
  author={Hershey, Shawn and Chaudhuri, Sourish and Ellis, Daniel PW and Gemmeke, Jort F and Jansen, Aren and Moore, R Channing and Plakal, Manoj and Platt, Devin and Saurous, Rif A and Seybold, Bryan and others},
  booktitle={2017 ieee international conference on acoustics, speech and signal processing (icassp)},
  pages={131--135},
  year={2017},
  organization={IEEE}
}

@inproceedings{singh2022flava,
  title={Flava: A foundational language and vision alignment model},
  author={Singh, Amanpreet and Hu, Ronghang and Goswami, Vedanuj and Couairon, Guillaume and Galuba, Wojciech and Rohrbach, Marcus and Kiela, Douwe},
  booktitle={Proceedings of the IEEE/CVF conference on computer vision and pattern recognition},
  pages={15638--15650},
  year={2022}
}

@inproceedings{gupta2016cross,
  title={Cross modal distillation for supervision transfer},
  author={Gupta, Saurabh and Hoffman, Judy and Malik, Jitendra},
  booktitle={Proceedings of the IEEE conference on computer vision and pattern recognition},
  pages={2827--2836},
  year={2016}
}

@article{vapnik2009new,
  title={A new learning paradigm: Learning using privileged information},
  author={Vapnik, Vladimir and Vashist, Akshay},
  journal={Neural networks},
  volume={22},
  number={5-6},
  pages={544--557},
  year={2009},
  publisher={Elsevier}
}

@inproceedings{hoffman2016learning,
  title={Learning with side information through modality hallucination},
  author={Hoffman, Judy and Gupta, Saurabh and Darrell, Trevor},
  booktitle={Proceedings of the IEEE conference on computer vision and pattern recognition},
  pages={826--834},
  year={2016}
}

@article{ofli2020analysis,
  title={Analysis of social media data using multimodal deep learning for disaster response},
  author={Ofli, Ferda and Alam, Firoj and Imran, Muhammad},
  journal={arXiv preprint arXiv:2004.11838},
  year={2020}
}
}

\newpage
\appendix
\section{Implementation Details}
\label{app:implementation_details}

\subsection{Computational Resources}
All experiments were conducted on a single workstation equipped with an AMD Ryzen 9 9900X CPU, an NVIDIA RTX PRO 6000 Blackwell Workstation Edition GPU with 96GB memory, and 64GB RAM. 
Unless otherwise specified, each training run was performed on a single GPU.

\begin{table}[h]
\centering
\caption{Implementation details for each dataset.}
\label{tab:training_details}
\small
\resizebox{\linewidth}{!}{
\begin{tabular}{lcccccccc}
\toprule
Dataset & Epochs & Learning rate & Weight decay & Batch size & \#Code &Code Dim. & $\beta_{\mathrm{com}}$ & $\lambda_{\mathrm{dist}}$\\
\midrule
RAVDESS       & 100 & $1\times10^{-3}$ & $1\times10^{-4}$ & 64  & 512 & 128 &0.25 & $1\times10^{-2}$\\
VGG-Sound      & 100 & $1\times10^{-3}$ & $1\times10^{-4}$ & 128 & 512 & 128  &0.25 & $1\times10^{-2}$\\
CrisisMMD     & 100 & $1\times10^{-3}$ & $1\times10^{-4}$ & 64  & 256 & 128 &0.25 & $1\times10^{-2}$\\
MM-IMDB       & 100 & $1\times10^{-3}$ & $1\times10^{-4}$ & 64  & 512 & 64 &0.25 & $1\times10^{-2}$\\
NYU-Depth V2  & 100 & $1\times10^{-4}$ & $1\times10^{-4}$ & 6   & 1024 & 256 &0.25 & $1\times10^{-2}$\\
\bottomrule
\end{tabular}
}
\end{table}

\subsection{General Training Details}
Table~\ref{tab:training_details} summarizes the training settings used in our experiments. 
Unless otherwise specified, we train each model using AdamW with $\beta=(0.9, 0.999)$. 
For a fair comparison, additional training stages required by FitNets~\cite{romero2015fitnetshintsdeepnets}, MGDFR~\cite{xue2023modality}, and our method are also trained for 100 epochs. 
These correspond to the hint regression stage of FitNets, the co-training stage of MGDFR, and Stage-1 of our method, respectively.

\subsection{Implementation Details for Distillation}
For our method, codebook sizes (\#Code), dimension of each code (Code Dim.), commitment coefficient $\beta_{\mathrm{com}}$  and distillation loss coefficient $\lambda_{\mathrm{dist}}$ are summarized in Table~\ref{tab:training_details}.
We use a linear projector to map the intermediate features into the code space before vector quantization.
Quantization is performed by assigning each projected feature to its nearest code according to the squared Euclidean distance.
During Stage-1, we additionally include a small entropy-based code usage regularization term to stabilize codebook learning.
Before Stage-3, we re-initialize the student parameters and train the final student from scratch using the selected teacher codebook.

For feature-alignment baselines, we use pooled features for feature-level distillation when the teacher and student intermediate features have different structural forms, and add a projection layer if necessary. When dense correspondence between teacher and student features is available, as in NYU-Depth V2, we apply feature-level distillation directly to the corresponding intermediate features without pooling.

\subsection{RAVDESS}
\paragraph{Dataset Specification.}
We use the Ryerson Audio-Visual Database of Emotional Speech and Song (RAVDESS) for audio-visual emotion classification~\cite{livingstone2018ryerson}. 
In our experiments, we use the speech subset and construct paired speech-audio and speech-video samples. 
The task is formulated as an 8-way emotion classification problem with the following emotion categories: \textit{neutral}, \textit{calm}, \textit{happy}, \textit{sad}, \textit{angry}, \textit{fearful}, \textit{disgust}, and \textit{surprised}. 
Each paired audio-video sample shares the same emotion label. 
The resulting dataset contains 1,440 paired samples, which are split into training, validation, and test sets with a ratio of 60\%/20\%/20\%.

\paragraph{Preprocessing.}
For audio data, we convert each waveform to mono when necessary, resample it to 16 kHz, and normalize its duration to 3.6 seconds. 
Longer clips are center-cropped, while shorter clips are symmetrically zero-padded. 
The resulting waveform is passed through a pretrained Wav2Vec2-Base model~\cite{baevski2020wav2vec}, and the final hidden representations are used as the audio feature sequence. 
The extracted audio features are arranged as $D \times T = 768\times 179$, where $D$ denotes the feature dimension and $T$ denotes the temporal length.
For video data, we uniformly sample 10 frames from each video. 
Each frame is center-cropped, resized to $224 \times 224$, and normalized using ImageNet statistics~\cite{deng2009imagenet}. 
We then feed each frame into an ImageNet-pretrained ResNet-18~\cite{he2016deep} and use the layer-3 feature map as the frame-level visual representation. 
The frame-level features are averaged over the sampled frames to obtain a single spatial feature map for each video. 
The extracted visual features have size $D \times H\times W = 256 \times 14 \times 14$, where $D$ denotes the channel (feature) dimension and $H\times W$ denotes the spatial resolution.

\paragraph{Model.}
We use lightweight classifiers on top of the pre-extracted audio and video features.
The audio classifier receives Wav2Vec2 features and applies two 1D convolutional blocks.
The resulting encoder output, with shape $B \times 256 \times T$, is used as the audio intermediate feature, where $B$ denotes batch size.
It is average-pooled over the temporal dimension and passed through a linear layer with ReLU activation and an 8-way classifier.
The video classifier receives ResNet-18 layer-3 features and applies two 2D convolutional blocks.
The resulting encoder output, with shape $B \times 256 \times H \times W$, is used as the video intermediate feature, where $B$ denotes batch size.
It is average-pooled over the spatial dimensions and passed through a linear layer with ReLU activation and an 8-way classifier.
In both classifiers, each convolutional block consists of convolution, batch normalization, and ReLU activation, and the hidden dimension before the final classifier is set to 64.

\subsection{VGG-Sound}
\paragraph{Dataset Specification.}
VGG-Sound~\cite{chen2020vggsound} is a large-scale audio-visual event classification dataset consisting of short audio-video clips collected from YouTube videos. 
Following~\cite{zhou2022contrastive,li2025mst}, we use a subset of 48,755 paired samples covering 141 real-world event classes. 
Each paired audio-video sample shares the same event label, and the task is formulated as 141-way event classification. 
We split the dataset into training, validation, and test sets with a ratio of 60\%/20\%/20\%.

\paragraph{Preprocessing.}
Following the preprocessing protocol of \cite{zhou2022contrastive}, the visual features are extracted using VGG19~\cite{simonyan2014very} and the audio features are extracted using VGGish~\cite{hershey2017cnn}. 
Each audio feature is represented as a temporal sequence of size $T \times D = 10 \times 128$, where $T$ denotes the temporal length and $D$ denotes the feature dimension. 
For visual features, the original pre-extracted representation contains frame-level spatial features. 
We average the visual features over the temporal dimension and use the resulting spatial feature map of size $H \times W \times D = 7 \times 7 \times 512$, where $H \times W$ denotes the spatial resolution and $D$ denotes the channel dimension. 

\paragraph{Model.}
We use lightweight modality-specific classifiers on top of the pre-extracted features.
The audio classifier applies two 1D convolutional blocks to obtain the audio intermediate feature of shape $B \times 128 \times T$, where $B$ denotes batch size.
This intermediate feature is further processed by an additional 1D convolutional stage, average-pooled over time, and passed through a linear layer with ReLU activation and a 141-way classifier.
The video classifier applies a $1\times1$ convolutional reduction layer to obtain the video intermediate feature of shape $B \times 128 \times H \times W$.
The intermediate feature is further processed by a 2D convolutional block, average-pooled over the spatial dimensions, and passed through a linear layer with ReLU activation and a classifier.
In both classifiers, convolutional blocks consist of convolution, batch normalization, and ReLU activation.
We set the hidden dimension before the final classifier to 128.

\subsection{CrisisMMD}
\paragraph{Dataset Specification.}
We use CrisisMMD-V2~\cite{ofli2020analysis}, an updated version of the CrisisMMD multimodal disaster dataset~\cite{alam2018crisismmd}, for image-text tweet classification in humanitarian crisis settings.
The dataset contains 16,058 paired image-text samples, and the task is formulated as 8-way humanitarian category classification.
The categories are \textit{not\_humanitarian}, \textit{affected\_individuals}, \textit{infrastructure\_and\_utility\_damage}, \textit{injured\_or\_dead\_people}, \textit{missing\_or\_found\_people}, \textit{rescue\_volunteering\_or\_donation\_effort}, \textit{vehicle\_damage}, and \textit{other\_relevant\_information}.
Each paired image-text sample shares the same label.
We split the dataset into training, validation, and test sets with a ratio of 60\%/20\%/20\%.

\paragraph{Preprocessing.}
For image data, we resize each image to $224 \times 224$ and normalize it using ImageNet statistics~\cite{deng2009imagenet}.
The processed image is passed through an ImageNet-pretrained ResNet-34~\cite{he2016deep}, and we use the layer-3 feature map as the image representation.
The extracted image features have size $D \times H \times W = 256 \times 14 \times 14$, where $D$ denotes the channel dimension and $H \times W$ denotes the spatial resolution.
For text data, we use BERTweet~\cite{nguyen2020bertweet} to extract token-level representations from the tweet text.
Each tweet is tokenized with a maximum length of 128, and the last hidden states of BERTweet are used as the text feature sequence.
The extracted text features have size $T_i \times D = T_i \times 768$, where $T_i$ denotes the valid token length of the $i$-th tweet and $D$ denotes the feature dimension.

\paragraph{Model.}
We use lightweight modality-specific classifiers on top of the pre-extracted image and text features.
The image classifier applies two 2D convolutional blocks.
The resulting encoder output, with shape $B \times 128 \times H \times W$, is used as the image intermediate feature, where $B$ denotes batch size.
It is average-pooled over the spatial dimensions and passed through a linear layer with ReLU activation and an 8-way classifier.
The text classifier applies two 1D convolutional blocks.
The resulting encoder output, with shape $B \times 128 \times T$, is used as the text intermediate feature, where $T$ denotes the token sequence length.
For text pooling, we average only over valid token positions using the token lengths.
The pooled representation is then passed through a linear layer with ReLU activation and an 8-way classifier.
In both classifiers, each convolutional block consists of convolution, batch normalization, and ReLU activation, with hidden dimensions set to 64.

\subsection{MM-IMDB}
\paragraph{Dataset Specification.}
We use MM-IMDB~\cite{arevalo2017gated} for image-text movie genre classification.
Although the original dataset is a 23-class multi-label genre classification benchmark, we construct a single-label subset for our experiments.
Several distillation methods in our comparison are formulated for single-label classification using softmax-normalized logits, and extending them to the multi-label setting would require redesigning their original distillation formulations.
We therefore adopt a controlled single-label setting so that the comparison primarily reflects differences among the CMKD methods rather than differences in their multi-label extensions.
Specifically, we first remove four extremely rare genres, \textit{News}, \textit{Adult}, \textit{Talk-Show}, and \textit{Reality-TV}, from the genre labels of each sample, and discard samples that have no remaining genre label.
We then retain only samples with exactly one remaining genre label.
Finally, we remove genres with fewer than 50 single-label samples, together with their samples.
The resulting dataset contains 5,778 paired poster-text samples from 8 classes: \textit{Action}, \textit{Comedy}, \textit{Documentary}, \textit{Drama}, \textit{Horror}, \textit{Sci-Fi}, \textit{Thriller}, and \textit{Western}, and the number of samples per class is reported in Table~\ref{tab:mmimdb_dist}.
Each paired poster-text sample shares the same label.
Because this filtering substantially changes both the size and the class distribution of the original benchmark, we do not use the official split.
Instead, for each seed, we construct a stratified 60\%/20\%/20\% split for training, validation, and testing, which is shared across all compared methods.
Accordingly, the MM-IMDB results should be interpreted as a customized single-label proxy for controlled CMKD comparison, and they are not directly comparable with results reported under the standard multi-label protocol.

\begin{table}[t]
\centering
\small
\caption{Number of samples per class in the single-label MM-IMDB dataset.}
\vspace{1mm}
\label{tab:mmimdb_dist}
\begin{tabular}{lr}
\toprule
Class & \# Samples\\
\midrule
Drama       & 2,681 \\
Comedy      & 1,337 \\
Documentary & 918 \\
Horror      & 417 \\
Western     & 173 \\
Thriller    & 143 \\
Sci-Fi      & 56 \\
Action      & 53 \\
\midrule
Total       & 5,778 \\
\bottomrule
\end{tabular}
\end{table}

\paragraph{Preprocessing.}
For image and text preprocessing, we use FLAVA~\cite{singh2022flava} to extract modality-specific hidden representations from movie posters and plot descriptions.
For image data, each poster is processed by the FLAVA image encoder, whose output consists of a CLS token and $14 \times 14$ patch tokens.
To reduce the sequence length, we average-pool the patch tokens over non-overlapping $2 \times 2$ regions while preserving the CLS token.
This yields image features of size $L \times D = 50 \times 768$, where $L=1+7\times7$ denotes the resulting token length and $D$ denotes the feature dimension.
For text data, we use the movie plot as the text input.
If the plot is unavailable, we use the movie title as a fallback.
The text is tokenized with a maximum length of 128 and passed through the FLAVA text encoder.
We preserve the CLS token and adaptively average-pool the remaining token representations to obtain a fixed-length text sequence.
This yields text features of size $L \times D = 64 \times 768$, where $L$ denotes the pooled token length and $D$ denotes the feature dimension.

\paragraph{Model.}
We use lightweight sequence classifiers on top of the pre-extracted FLAVA image and text features. For both modalities, each token feature is passed through a linear projection, layer normalization, and GELU activation to obtain the intermediate feature for distillation. The intermediate feature has shape $B \times 50 \times 64$ for image data and $B \times 64 \times 64$ for text data. We use attention pooling to obtain the pooled representation, with the text attention mask used to ignore padded positions. The pooled representation is passed through layer normalization, a linear layer with GELU activation, and a final 8-way classifier. Hidden dimensions are set to 64.

\subsection{NYU-Depth V2}
\paragraph{Dataset Specification.}
We use NYU-Depth V2~\cite{silberman2012indoor} for RGB-depth semantic segmentation for indoor scenes.
The dataset contains 1,449 paired RGB-depth samples captured from indoor scenes, with pixel-wise label maps of size $480 \times 640$.
The semantic categories include common indoor structures and objects, such as \textit{wall}, \textit{floor}, \textit{ceiling}, \textit{bed}, \textit{chair}, \textit{table}, \textit{door}, and \textit{window}.
We use 41 label indices, consisting of 40 semantic classes and one void class, and exclude the void class from evaluation.
We split the dataset into training, validation, and test sets with a ratio of 60\%/20\%/20\%.

\paragraph{Preprocessing.}
For RGB data, we first apply min-max normalization to each image, resize it to $224 \times 224$, and normalize it using ImageNet statistics~\cite{deng2009imagenet}.
The processed RGB image is passed through an ImageNet-pretrained ResNet-34~\cite{he2016deep}, and we use the layer-3 feature map as the RGB representation.
For depth data, we apply min-max normalization to each depth map and replicate the single-channel depth input into three channels.
We then apply the same $224 \times 224$ resizing and ImageNet normalization, and pass the result through the same ResNet-34 layer-3 feature extractor.
The extracted RGB and depth features have size $D \times H \times W = 256 \times 14 \times 14$, where $D$ denotes the channel dimension and $H \times W$ denotes the spatial resolution.
The semantic label maps are kept at their original spatial resolution of $480 \times 640$ and are used as pixel-wise supervision for segmentation.

\paragraph{Model.}
We use lightweight segmentation heads on top of the pre-extracted RGB and depth features.
The RGB and depth branches share the same architecture and differ only in the input modality.
Each branch applies two 2D convolutional blocks to the ResNet-34 layer-3 feature map.
The resulting encoder output, with shape $B \times 256 \times 14 \times 14$, is used as the intermediate feature for distillation, where $B$ denotes batch size.
For segmentation prediction, the intermediate feature is passed through an upsampling decoder.
The decoder progressively upsamples the feature map and produces dense semantic logits with spatial size $480 \times 640$.
The output dimension is 41, corresponding to 40 semantic classes and one void class.
The model is trained with pixel-wise cross-entropy loss, where the void label is ignored.
For NYU-Depth V2, where RGB and depth features are spatially aligned, we apply each feature-alignment baseline to the corresponding dense intermediate feature maps whenever possible, rather than pooling the features.

\section{Additional Experimental Results}
\subsection{Results with Standard Deviations}
Tables~\ref{tab:cls_main_std} and~\ref{tab:seg_main_std} report the classification and semantic segmentation results from the main paper with standard deviations over five independent runs. 
The mean values are identical to those reported in the corresponding main result tables, while the standard deviations provide additional information about variability across random seeds.
Tables~\ref{tab:loss_ablation_std}, \ref{tab:pred_matching_target_std}, and~\ref{tab:score_ablation_std} further report the ablation results in Tables~\ref{tab:loss_ablation}, \ref{tab:pred_matching_target}, and~\ref{tab:score_ablation} with standard deviations under the same protocol.

\begin{table*}[!t]
\small
\caption{
Classification results. 
A, V, T, and I denote audio, video, text, and image modalities, respectively, and each arrow indicates the teacher-to-student distillation direction. 
We report mean accuracy $\pm$ standard deviation over five independent runs, with the best mean result in each column shown in bold.
}
\vspace{1mm}
\label{tab:cls_main_std}
\centering
\begin{tabular}{l|cccc}
\toprule
\multirow{2}{*}{Method}& \multicolumn{2}{c}{RAVDESS} & \multicolumn{2}{c}{VGG-Sound}\\
 & A $\rightarrow$ V & V $\rightarrow$ A & A $\rightarrow$ V & V $\rightarrow$ A\\
\midrule
w/o KD  &0.9208$\pm$0.0164&0.6826$\pm$0.0189  &0.5031$\pm$0.0035&0.6058$\pm$0.0086\\
\midrule
KLD     &0.9118$\pm$0.0156&0.6972$\pm$0.0674  &0.5153$\pm$0.0067&0.6176$\pm$0.0090\\
MLLD    &0.9292$\pm$0.0140&0.6889$\pm$0.0461  &0.5135$\pm$0.0056&0.6122$\pm$0.0047\\
FitNets &0.9062$\pm$0.0188&0.7146$\pm$0.0385  &0.4916$\pm$0.0105&0.5962$\pm$0.0128\\
RKD     &0.9236$\pm$0.0132&0.7014$\pm$0.0269  &0.5066$\pm$0.0068&0.6085$\pm$0.0018\\
CRD     &0.9299$\pm$0.0199&0.7007$\pm$0.0246  &0.5063$\pm$0.0032&0.6110$\pm$0.0075\\
OFA     &0.9215$\pm$0.0183&0.6562$\pm$0.0577  &0.5100$\pm$0.0093&0.6129$\pm$0.0050\\
\midrule
MGDFR   &0.9313$\pm$0.0206&0.7264$\pm$0.0090  &0.5225$\pm$0.0034&0.6186$\pm$0.0084\\
C2KD    &0.9306$\pm$0.0139&0.7021$\pm$0.0314  &0.5123$\pm$0.0049&0.6154$\pm$0.0045\\
STXD    &0.9035$\pm$0.0093&0.7250$\pm$0.0165  &0.5067$\pm$0.0048&0.6093$\pm$0.0033\\
FDD     &0.9153$\pm$0.0090&0.7055$\pm$0.0158  &0.5028$\pm$0.0078&0.6063$\pm$0.0093\\
\midrule
Ours    &\textbf{0.9431}$\pm$0.0097&\textbf{0.7486}$\pm$0.0231  &\textbf{0.5299}$\pm$0.0017&\textbf{0.6229}$\pm$0.0059 \\ 
\bottomrule
\end{tabular}
\par\vspace{3mm}
\begin{tabular}{l|cccc}
\toprule
\multirow{2}{*}{Method}& \multicolumn{2}{c}{CrisisMMD} & \multicolumn{2}{c}{MM-IMDB}\\
 & T $\rightarrow$ I & I $\rightarrow$ T & T $\rightarrow$ I & I $\rightarrow$ T\\
\midrule
w/o KD  &0.5403$\pm$0.0085&0.5500$\pm$0.0050 &0.6645$\pm$0.0092&0.7294$\pm$0.0159\\
\midrule
KLD     &0.5377$\pm$0.0063&0.5578$\pm$0.0049 &0.6732$\pm$0.0103&0.7310$\pm$0.0079\\
MLLD    &0.5357$\pm$0.0062&0.5562$\pm$0.0059 &0.6699$\pm$0.0113&0.7320$\pm$0.0096\\
FitNets &0.5281$\pm$0.0202&0.5507$\pm$0.0091 &0.6759$\pm$0.0101&0.7296$\pm$0.0114\\
RKD     &0.5432$\pm$0.0058&0.5569$\pm$0.0061 &0.6668$\pm$0.0117&0.7317$\pm$0.0109\\
CRD     &0.5377$\pm$0.0110&0.5560$\pm$0.0114 &0.6685$\pm$0.0109&0.7299$\pm$0.0077\\
OFA     &0.5362$\pm$0.0080&0.5565$\pm$0.0032 &0.6715$\pm$0.0150&0.7339$\pm$0.0105\\
\midrule
MGDFR   &\textbf{0.5440}$\pm$0.0103&0.5584$\pm$0.0026 &0.6725$\pm$0.0105&0.7382$\pm$0.0135\\
C2KD    &0.5410$\pm$0.0052&0.5520$\pm$0.0082 &0.6670$\pm$0.0041&0.7372$\pm$0.0101\\
STXD    &0.5421$\pm$0.0069&0.5485$\pm$0.0083 &0.6761$\pm$0.0116&0.7265$\pm$0.0168\\
FDD     &0.5382$\pm$0.0088&0.5547$\pm$0.0067 &0.6670$\pm$0.0108&0.7323$\pm$0.0105\\
\midrule
Ours    &0.5433$\pm$0.0062&\textbf{0.5592}$\pm$0.0071 &\textbf{0.6768}$\pm$0.0055&\textbf{0.7434}$\pm$0.0139\\ 
\bottomrule
\end{tabular}
\end{table*}
			
\begin{table}[t]
\centering
\small
\caption{Semantic segmentation results on NYU-Depth. Quantitative comparison for two distillation scenarios, where the first modality denotes the teacher and the second denotes the student. Results are reported as mean $\pm$ standard deviation over five independent runs.}
\resizebox{\linewidth}{!}{
\begin{tabular}{l|cccccc}
\toprule
\multirow{2}{*}{Method} & \multicolumn{3}{c}{Depth $\rightarrow$ RGB} & \multicolumn{3}{c}{RGB $\rightarrow$ Depth}  \\
& OA $\uparrow$ & CA $\uparrow$ & mIoU $\uparrow$ & OA $\uparrow$ & CA $\uparrow$ & mIoU $\uparrow$\\
\midrule
w/o KD  &0.5582$\pm$0.0112&0.2748$\pm$0.0218&0.1850$\pm$0.0154&\textbf{0.5269}$\pm$0.0070&0.2100$\pm$0.0155&0.1366$\pm$0.0110 \\
\midrule
MLLD    &0.5599$\pm$0.0122&0.2792$\pm$0.0226&0.1911$\pm$0.0169&0.5250$\pm$0.0063&0.2118$\pm$0.0150&0.1399$\pm$0.0152\\
CRD     &0.5592$\pm$0.0101&0.2677$\pm$0.0158&0.1806$\pm$0.0108&0.5261$\pm$0.0061&0.2111$\pm$0.0125&0.1385$\pm$0.0107\\
OFA     &0.5607$\pm$0.0119&0.2881$\pm$0.0146&0.1982$\pm$0.0107&0.5242$\pm$0.0085&0.2123$\pm$0.0125&0.1409$\pm$0.0123\\
MGDFR   &0.5601$\pm$0.0123&0.2737$\pm$0.0196&0.1878$\pm$0.0165&0.5213$\pm$0.0099&0.2162$\pm$0.0097&0.1442$\pm$0.0083\\
C2KD	&0.5587$\pm$0.0089&0.2771$\pm$0.0090&0.1878$\pm$0.0052&0.5214$\pm$0.0062&0.2093$\pm$0.0092&0.1396$\pm$0.0074\\
STXD    &0.5616$\pm$0.0108&0.2786$\pm$0.0220&0.1925$\pm$0.0171&0.5231$\pm$0.0102&0.2164$\pm$0.0125&0.1425$\pm$0.0124\\
FDD     &0.5608$\pm$0.0114&0.2744$\pm$0.0279&0.1855$\pm$0.0234&0.5252$\pm$0.0063&0.2122$\pm$0.0146&0.1403$\pm$0.0153\\
\midrule
Ours     &\textbf{0.5632}$\pm$0.0103&\textbf{0.2909}$\pm$0.0260&\textbf{0.2003}$\pm$0.0194&0.5218$\pm$0.0078&\textbf{0.2184}$\pm$0.0154&\textbf{0.1456}$\pm$0.0105\\
\bottomrule
\end{tabular}
}
\label{tab:seg_main_std}
\end{table}

\begin{table}[t]
\centering
\small
\caption{Ablation study on Stage-3 loss components with standard deviations.
``Code'' denotes the student-side codebook-branch objective.
``Pred-Match'' denotes the codebook-induced prediction matching term.
Results are reported as mean $\pm$ standard deviation over five independent runs.}
\vspace{1mm}
\label{tab:loss_ablation_std}
\begin{tabular}{cc|cccc}
\toprule
\multirow{2}{*}{Code} & \multirow{2}{*}{Pred-Match} & \multicolumn{2}{c}{RAVDESS} & \multicolumn{2}{c}{VGG-Sound}\\
& & A $\rightarrow$ V & V $\rightarrow$ A & A $\rightarrow$ V & V $\rightarrow$ A\\
\midrule
- & - & 0.9208$\pm$0.0164 & 0.6826$\pm$0.0189 & 0.5031$\pm$0.0035 & 0.6058$\pm$0.0086 \\
\checkmark & - & 0.9202$\pm$0.0260 & 0.7333$\pm$0.0397 & 0.5067$\pm$0.0029 & 0.6098$\pm$0.0033 \\
- & \checkmark & 0.9285$\pm$0.0234 & 0.7479$\pm$0.0175 & 0.5294$\pm$0.0013 & 0.6204$\pm$0.0056 \\
\checkmark & \checkmark & 0.9431$\pm$0.0097 & 0.7486$\pm$0.0231 & 0.5299$\pm$0.0017 & 0.6229$\pm$0.0059 \\
\bottomrule
\end{tabular}
\par\vspace{3mm}
\begin{tabular}{cc|cccc}
\toprule
\multirow{2}{*}{Code} & \multirow{2}{*}{Pred-Match} & \multicolumn{2}{c}{CrisisMMD} & \multicolumn{2}{c}{MM-IMDB}\\
& & T $\rightarrow$ I & I $\rightarrow$ T & T $\rightarrow$ I & I $\rightarrow$ T\\
\midrule
- & - & 0.5403$\pm$0.0085 & 0.5500$\pm$0.0050 & 0.6645$\pm$0.0092 & 0.7294$\pm$0.0159 \\
\checkmark & - & 0.5387$\pm$0.0129 & 0.5534$\pm$0.0037 & 0.6716$\pm$0.0037 & 0.7363$\pm$0.0065 \\
- & \checkmark & 0.5367$\pm$0.0100 & 0.5529$\pm$0.0018 & 0.6770$\pm$0.0054 & 0.7421$\pm$0.0136 \\
\checkmark & \checkmark & 0.5433$\pm$0.0062 & 0.5592$\pm$0.0071 & 0.6768$\pm$0.0055 & 0.7434$\pm$0.0139 \\
\bottomrule
\end{tabular}
\end{table}

\begin{table}[t]
\centering
\small
\caption{Ablation on the prediction matching target in Stage-3 loss with standard deviations.
Results are reported as mean $\pm$ standard deviation over five independent runs.}
\vspace{1mm}
\label{tab:pred_matching_target_std}
\begin{tabular}{l|cccc}
\toprule
\multirow{2}{*}{Pred-Match Target} & \multicolumn{2}{c}{RAVDESS} & \multicolumn{2}{c}{VGG-Sound}\\
& A $\rightarrow$ V & V $\rightarrow$ A & A $\rightarrow$ V & V $\rightarrow$ A\\
\midrule
Original branch $(\mathbf{p}^{(t)})$ & 0.8903$\pm$0.0140 & 0.6201$\pm$0.0788 & 0.5173$\pm$0.0024 & 0.6171$\pm$0.0079 \\
Codebook branch $(\mathbf{q}^{(t)})$ & 0.9431$\pm$0.0097 & 0.7486$\pm$0.0231 & 0.5299$\pm$0.0017 & 0.6229$\pm$0.0059 \\
\bottomrule
\end{tabular}
\par\vspace{3mm}
\begin{tabular}{l|cccc}
\toprule
\multirow{2}{*}{Pred-Match Target} & \multicolumn{2}{c}{CrisisMMD} & \multicolumn{2}{c}{MM-IMDB}\\
& T $\rightarrow$ I & I $\rightarrow$ T & T $\rightarrow$ I & I $\rightarrow$ T\\
\midrule
Original branch $(\mathbf{p}^{(t)})$ & 0.5425$\pm$0.0081 & 0.5539$\pm$0.0046 & 0.6654$\pm$0.0158 & 0.7351$\pm$0.0139 \\
Codebook branch $(\mathbf{q}^{(t)})$ & 0.5433$\pm$0.0062 & 0.5592$\pm$0.0071 & 0.6768$\pm$0.0055 & 0.7434$\pm$0.0139 \\
\bottomrule
\end{tabular}
\end{table}

\begin{table}[t]
\centering
\small
\caption{Ablation study on code selection criteria with standard deviations.
Results are reported as mean $\pm$ standard deviation over five independent runs.}
\vspace{1mm}
\label{tab:score_ablation_std}
\begin{tabular}{l|cccc}
\toprule
\multirow{2}{*}{Selection Criteria} & \multicolumn{2}{c}{RAVDESS} & \multicolumn{2}{c}{VGG-Sound}\\
& A $\rightarrow$ V & V $\rightarrow$ A & A $\rightarrow$ V & V $\rightarrow$ A\\
\midrule
Random & 0.9292$\pm$0.0134 & 0.7396$\pm$0.0214 & 0.5252$\pm$0.0049 & 0.6211$\pm$0.0051 \\
Task relevance only & 0.9396$\pm$0.0063 & 0.7361$\pm$0.0134 & 0.5276$\pm$0.0055 & 0.6233$\pm$0.0063 \\
Student compatibility only & 0.9375$\pm$0.0078 & 0.7417$\pm$0.0124 & 0.5275$\pm$0.0077 & 0.6222$\pm$0.0052 \\
Combination (Ours) & 0.9431$\pm$0.0097 & 0.7486$\pm$0.0231 & 0.5299$\pm$0.0017 & 0.6229$\pm$0.0059 \\
\bottomrule
\end{tabular}
\par\vspace{3mm}
\begin{tabular}{l|cccc}
\toprule
\multirow{2}{*}{Selection Criteria} & \multicolumn{2}{c}{CrisisMMD} & \multicolumn{2}{c}{MM-IMDB}\\
& T $\rightarrow$ I & I $\rightarrow$ T & T $\rightarrow$ I & I $\rightarrow$ T\\
\midrule
Random & 0.5353$\pm$0.0076 & 0.5536$\pm$0.0043 & 0.6718$\pm$0.0135 & 0.7353$\pm$0.0116 \\
Task relevance only & 0.5435$\pm$0.0047 & 0.5558$\pm$0.0072 & 0.6727$\pm$0.0139 & 0.7351$\pm$0.0104 \\
Student compatibility only & 0.5437$\pm$0.0052 & 0.5565$\pm$0.0058 & 0.6722$\pm$0.0138 & 0.7346$\pm$0.0103 \\
Combination (Ours) & 0.5433$\pm$0.0062 & 0.5592$\pm$0.0071 & 0.6768$\pm$0.0055 & 0.7434$\pm$0.0139 \\
\bottomrule
\end{tabular}
\end{table}

\subsection{Statistical Significance Analysis}
\label{app:significance}

Following the protocol of MST-Distill~\cite{li2025mst}, the data split and the teacher checkpoint vary across seeds in our main experiments.
Within each seed, all methods share the same data split and teacher checkpoint, ensuring a fair comparison under identical experimental conditions.
While this protocol evaluates performance across diverse CMKD settings, the reported standard deviations include variability arising from the data split and the teacher model in addition to that arising from the distillation method itself.

\paragraph{Controlled Significance Test.}
To isolate the effect of the distillation method, we conduct controlled experiments in which the data split and the teacher checkpoint are fixed and only the remaining training randomness is varied over ten runs.
We compare our method with the strongest competing method in Table~\ref{tab:cls_result} using a two-sided paired t-test.
As shown in Table~\ref{tab:sig_main}, our method achieves statistically significant improvements ($p<0.05$) in five of the eight transfer directions.
Two additional settings, CrisisMMD I$\rightarrow$T and MM-IMDB T$\rightarrow$I, show near-threshold evidence with p-values close to 0.05.
In CrisisMMD T$\rightarrow$I, MGDFR achieves a slightly higher mean, but the evidence for a reliable difference is limited, indicating that our method remains competitive in this setting.

\paragraph{Average Rank Analysis.}
We further evaluate whether the performance gains remain consistent across variations in the data split under the original varying-split protocol.
For each seed, we rank all methods evaluated with the same data split and teacher checkpoint, and then average the ranks over five seeds.
As shown in Table~\ref{tab:avg_rank}, our method achieves the best average rank in seven of the eight transfer directions and the second-best average rank in CrisisMMD T$\rightarrow$I.
This indicates that the proposed method remains consistently competitive across changes in both the data split and the teacher checkpoint, rather than benefiting from only a few favorable runs.

\paragraph{Significance of Loss Components.}
Under the same controlled setting, we examine whether the contribution of the ``Code'' term can be explained by run-to-run noise.
As shown in Table~\ref{tab:sig_loss}, adding the ``Code'' term to ``Pred-Match'' improves the mean performance in all eight transfer directions.
The improvement is statistically significant in four settings, with suggestive trends in two additional settings.
``Pred-Match'' is expected to produce a larger individual gain because it provides supervision more directly related to the task output, whereas the ``Code'' term regularizes the intermediate representation.
Note that the target of ``Pred-Match'' is the codebook-induced teacher prediction $\mathbf{q}^{(t)}$ rather than the original teacher prediction $\mathbf{p}^{(t)}$.
Both terms are therefore derived from the learned teacher codebook and provide complementary supervision at the representation and prediction levels.

\paragraph{Significance of Code Selection Criterion.}
We further evaluate whether the proposed code selection criterion provides a reliable advantage over random code selection.
As shown in Table~\ref{tab:sig_score}, the combined criterion improves the mean performance in all eight transfer directions and achieves statistically significant improvements in seven of them.
The remaining MM-IMDB I$\rightarrow$T setting also shows a positive trend ($p=0.0659$).
This indicates that the proposed criterion identifies codes that are systematically more useful for cross-modal distillation than random selection.

\begin{table}[t]
\centering
\small
\caption{Controlled significance tests against the runner-up methods for the classification results.
The data split and the teacher checkpoint are fixed, and results are reported as mean $\pm$ standard deviation over ten runs.
p-values are computed using a two-sided paired t-test.
The runner-up is STXD for MM-IMDB T$\rightarrow$I and MGDFR for all other settings.}
\vspace{1mm}
\label{tab:sig_main}
\begin{tabular}{l|cccc}
\toprule
\multirow{2}{*}{Method} & \multicolumn{2}{c}{RAVDESS} & \multicolumn{2}{c}{VGG-Sound}\\
& A $\rightarrow$ V & V $\rightarrow$ A & A $\rightarrow$ V & V $\rightarrow$ A\\
\midrule
Runner-up & 0.9271$\pm$0.0080 & 0.7351$\pm$0.0144 & 0.5219$\pm$0.0046 & 0.6126$\pm$0.0043 \\
Ours      & 0.9420$\pm$0.0071 & 0.7556$\pm$0.0120 & 0.5273$\pm$0.0041 & 0.6158$\pm$0.0029 \\
\midrule
p-value   & 0.001 & 0.007 & 0.028 & 0.007 \\
\bottomrule
\end{tabular}
\par\vspace{3mm}
\begin{tabular}{l|cccc}
\toprule
\multirow{2}{*}{Method} & \multicolumn{2}{c}{CrisisMMD} & \multicolumn{2}{c}{MM-IMDB}\\
& T $\rightarrow$ I & I $\rightarrow$ T & T $\rightarrow$ I & I $\rightarrow$ T\\
\midrule
Runner-up & 0.5486$\pm$0.0062 & 0.5550$\pm$0.0034 & 0.6680$\pm$0.0048 & 0.7443$\pm$0.0057 \\
Ours      & 0.5448$\pm$0.0084 & 0.5581$\pm$0.0019 & 0.6724$\pm$0.0037 & 0.7482$\pm$0.0051 \\
\midrule
p-value   & 0.105 & 0.060 & 0.055 & 0.046 \\
\bottomrule
\end{tabular}
\end{table}

\begin{table}[t]
\centering
\small
\caption{Average rank of classification methods across five runs with varying data splits and teacher checkpoints.
Lower values indicate better performance, and the best average rank in each column is shown in bold.}
\vspace{1mm}
\label{tab:avg_rank}
\begin{tabular}{l|cccccccc}
\toprule
\multirow{2}{*}{Method} & \multicolumn{2}{c}{RAVDESS} & \multicolumn{2}{c}{VGG-Sound} & \multicolumn{2}{c}{CrisisMMD} & \multicolumn{2}{c}{MM-IMDB}\\
& A $\rightarrow$ V & V $\rightarrow$ A & A $\rightarrow$ V & V $\rightarrow$ A & T $\rightarrow$ I & I $\rightarrow$ T & T $\rightarrow$ I & I $\rightarrow$ T\\
\midrule
KLD     & 7.4 & 6.4 & 4.6 & 3.8 & 7.4 & 5.4 & \textbf{4.0} & 7.0 \\
MLLD    & 4.8 & 7.2 & 4.4 & 5.8 & 8.6 & 5.0 & 5.8 & 6.4 \\
FitNets & 7.8 & 5.8 & 10.8 & 10.8 & 7.4 & 7.8 & 4.4 & 7.4 \\
RKD     & 5.4 & 7.0 & 8.0 & 8.2 & 4.4 & 4.6 & 7.6 & 6.6 \\
CRD     & 4.6 & 7.4 & 7.8 & 7.8 & 6.2 & 4.8 & 5.8 & 7.0 \\
OFA     & 5.8 & 8.8 & 5.6 & 5.0 & 7.2 & 5.2 & 6.0 & 5.6 \\
\midrule
MGDFR   & 4.0 & 4.0 & 2.0 & 2.8 & \textbf{3.4} & 3.8 & 5.8 & 3.6 \\
C2KD    & 4.4 & 6.6 & 4.8 & 4.2 & 5.2 & 7.6 & 8.0 & 3.6 \\
STXD    & 10.2 & 3.8 & 7.8 & 7.8 & 5.0 & 9.8 & 4.6 & 8.8 \\
FDD     & 7.0 & 5.8 & 9.0 & 8.0 & 6.4 & 7.2 & 8.4 & 6.4 \\
\midrule
Ours    & \textbf{2.4} & \textbf{2.0} & \textbf{1.0} & \textbf{1.4} & 3.8 & \textbf{3.4} & \textbf{4.0} & \textbf{2.0} \\
\bottomrule
\end{tabular}
\end{table}

\begin{table}[t]
\centering
\small
\caption{Controlled significance tests for the contribution of the ``Code'' term.
The data split and the teacher checkpoint are fixed, and results are reported as mean $\pm$ standard deviation over ten runs.
p-values are computed using a two-sided paired t-test.}
\vspace{1mm}
\label{tab:sig_loss}
\begin{tabular}{cc|cccc}
\toprule
\multirow{2}{*}{Code} & \multirow{2}{*}{Pred-Match} & \multicolumn{2}{c}{RAVDESS} & \multicolumn{2}{c}{VGG-Sound}\\
& & A $\rightarrow$ V & V $\rightarrow$ A & A $\rightarrow$ V & V $\rightarrow$ A\\
\midrule
- & \checkmark & 0.9274$\pm$0.0110 & 0.7469$\pm$0.0060 & 0.5246$\pm$0.0031 & 0.6129$\pm$0.0031 \\
\checkmark & \checkmark & 0.9420$\pm$0.0071 & 0.7556$\pm$0.0120 & 0.5273$\pm$0.0041 & 0.6158$\pm$0.0029 \\
\midrule
\multicolumn{2}{c|}{p-value} & 0.0031 & 0.0885 & 0.1215 & 0.0504 \\
\bottomrule
\end{tabular}
\par\vspace{3mm}
\begin{tabular}{cc|cccc}
\toprule
\multirow{2}{*}{Code} & \multirow{2}{*}{Pred-Match} & \multicolumn{2}{c}{CrisisMMD} & \multicolumn{2}{c}{MM-IMDB}\\
& & T $\rightarrow$ I & I $\rightarrow$ T & T $\rightarrow$ I & I $\rightarrow$ T\\
\midrule
- & \checkmark & 0.5377$\pm$0.0057 & 0.5518$\pm$0.0081 & 0.6690$\pm$0.0045 & 0.7448$\pm$0.0028 \\
\checkmark & \checkmark & 0.5448$\pm$0.0084 & 0.5581$\pm$0.0019 & 0.6724$\pm$0.0037 & 0.7482$\pm$0.0051 \\
\midrule
\multicolumn{2}{c|}{p-value} & 0.0223 & 0.0348 & 0.0116 & 0.1302 \\
\bottomrule
\end{tabular}
\end{table}

\begin{table}[t]
\centering
\small
\caption{Controlled significance tests for the code selection criterion.
The data split and the teacher checkpoint are fixed, and results are reported as mean $\pm$ standard deviation over ten runs.
p-values are computed using a two-sided paired t-test.}
\vspace{1mm}
\label{tab:sig_score}
\begin{tabular}{l|cccc}
\toprule
\multirow{2}{*}{Selection Criteria} & \multicolumn{2}{c}{RAVDESS} & \multicolumn{2}{c}{VGG-Sound}\\
& A $\rightarrow$ V & V $\rightarrow$ A & A $\rightarrow$ V & V $\rightarrow$ A\\
\midrule
Random & 0.9316$\pm$0.0100 & 0.7423$\pm$0.0073 & 0.5242$\pm$0.0029 & 0.6125$\pm$0.0023 \\
Combination (Ours) & 0.9420$\pm$0.0071 & 0.7556$\pm$0.0120 & 0.5273$\pm$0.0041 & 0.6158$\pm$0.0029 \\
\midrule
p-value & 0.0333 & 0.0205 & 0.0309 & 0.0287 \\
\bottomrule
\end{tabular}
\par\vspace{3mm}
\begin{tabular}{l|cccc}
\toprule
\multirow{2}{*}{Selection Criteria} & \multicolumn{2}{c}{CrisisMMD} & \multicolumn{2}{c}{MM-IMDB}\\
& T $\rightarrow$ I & I $\rightarrow$ T & T $\rightarrow$ I & I $\rightarrow$ T\\
\midrule
Random & 0.5339$\pm$0.0074 & 0.5527$\pm$0.0061 & 0.6668$\pm$0.0043 & 0.7443$\pm$0.0024 \\
Combination (Ours) & 0.5448$\pm$0.0084 & 0.5581$\pm$0.0019 & 0.6724$\pm$0.0037 & 0.7482$\pm$0.0051 \\
\midrule
p-value & 0.0105 & 0.0147 & 0.0034 & 0.0659 \\
\bottomrule
\end{tabular}
\end{table}

\subsection{Computational Cost Analysis}
\label{app:comp_cost}
We analyze the training cost of the proposed method on VGG-Sound in terms of end-to-end wall-clock training time, peak GPU memory usage, and floating-point operations (FLOPs).
For FLOPs, we exclude the teacher and student backbone forward passes, which are largely shared across methods, and report only the additional scoring or alignment operations introduced by each method.
As shown in Table~\ref{tab:comp_cost}, our method requires the longest wall-clock training time because of its three-stage training procedure.
Other multi-stage or computationally intensive methods, such as CRD, FitNets, and MGDFR, also require substantially longer training time than the student trained without distillation.
However, the peak memory usage and alignment-related FLOPs of our method are not consistently higher than those of the compared methods.
For instance, our method requires less peak memory and fewer alignment-related FLOPs than MGDFR.
The leave-one-code-out scoring in Stage-2 accounts for only 7.9\% and 7.7\% of the total training time for V$\rightarrow$A and A$\rightarrow$V, respectively, indicating that it is not the dominant computational cost under the default configuration.

\paragraph{Effect of Codebook Size.}
Table~\ref{tab:codebook_size} reports the stage-wise training time and accuracy under different codebook sizes $B$.
The runtime of Stage-2 increases approximately linearly with $B$, whereas the runtimes of Stage-1 and Stage-3 remain nearly constant.
Performance is generally robust to $B$, and an intermediate codebook size ($B=512$, our default) provides the best trade-off between accuracy and computational cost.
We also observe that code utilization decreases for oversized codebooks.
With $B=2048$, more than 30\% of the codes frequently remain unused, and $B=1024$ yields approximately 10-15\% unused codes, whereas $B\leq512$ generally maintains 99-100\% code utilization.
This suggests that excessively large codebooks increase the scoring cost while introducing redundant or underutilized codes without consistent performance gains.
The code utilization measured in Stage-1 can therefore serve as a practical diagnostic for identifying unnecessarily large codebooks.
Nevertheless, since the cost of Stage-2 scales linearly with $B$, it may become a bottleneck for substantially larger codebooks, and more efficient importance estimation remains a direction for future work.

\paragraph{Inference Cost.}
All additional components, including the teacher, the codebook, the projection layers, and the codebook branches, as well as the Stage-2 scoring procedure, are used only during training.
At inference time, they are entirely removed, and only the original student encoder and task head are retained.
Therefore, the proposed method introduces no additional parameters, memory, or computation at inference compared with the student trained without distillation, and the teacher modality is not required at deployment.

\begin{table}[t]
\centering
\small
\caption{Training computational costs on VGG-Sound.
Time and Mem denote the end-to-end wall-clock training time (s) and peak GPU memory usage (MB), respectively.
TFLOPs are measured only for the method-specific scoring and alignment operations, excluding the teacher and student forward passes that are largely shared across methods.
$\hookrightarrow$ Stage-2 denotes the cost of the leave-one-code-out scoring in our method.}
\vspace{1mm}
\label{tab:comp_cost}
\begin{tabular}{l|cccccc}
\toprule
\multirow{2}{*}{Method} & \multicolumn{3}{c}{V $\rightarrow$ A} & \multicolumn{3}{c}{A $\rightarrow$ V}\\
& Time & Mem & TFLOPs & Time & Mem & TFLOPs\\
\midrule
w/o KD  & 82.69  & 25.40  & --    & 168.44 & 77.25  & --     \\
\midrule
KLD     & 172.95 & 76.29  & 0.62  & 176.04 & 80.34  & 0.62   \\
MSE     & 171.78 & 76.35  & 0.62  & 165.68 & 80.40  & 0.62   \\
CRD     & 325.45 & 88.22  & 0.63  & 324.03 & 94.71  & 0.63   \\
FitNets & 338.17 & 78.26  & 1.24  & 339.19 & 83.05  & 1.24   \\
MGDFR   & 327.07 & 178.07 & 76.81 & 316.99 & 177.95 & 784.21 \\
\midrule
Ours    & 458.85 & 122.86 & 31.56 & 452.29 & 103.65 & 148.21 \\
\quad$\hookrightarrow$ Stage-2 & 36.24 & 95.69 & 31.56 & 34.87 & 93.18 & 148.21 \\
\bottomrule
\end{tabular}
\end{table}

\begin{table}[t]
\centering
\small
\caption{Stage-wise training time (s) and accuracy under different codebook sizes $B$ on VGG-Sound.
Accuracy is averaged over five independent runs.
Our default setting is $B=512$.}
\vspace{1mm}
\label{tab:codebook_size}
\begin{tabular}{c|cccc|cccc}
\toprule
\multirow{2}{*}{$B$} & \multicolumn{4}{c|}{V $\rightarrow$ A} & \multicolumn{4}{c}{A $\rightarrow$ V}\\
& Stage-1 & Stage-2 & Stage-3 & Acc. & Stage-1 & Stage-2 & Stage-3 & Acc.\\
\midrule
128  & 231.53 & 10.48  & 182.34 & 0.6202 & 224.55 & 10.56  & 179.26 & 0.5270 \\
256  & 228.37 & 18.54  & 190.97 & 0.6213 & 225.65 & 18.65  & 193.02 & 0.5272 \\
512  & 226.81 & 36.24  & 195.80 & 0.6229 & 225.70 & 34.87  & 191.72 & 0.5299 \\
1024 & 226.90 & 79.93  & 191.29 & 0.6201 & 224.92 & 68.62  & 190.45 & 0.5283 \\
2048 & 228.29 & 184.83 & 202.42 & 0.6205 & 221.61 & 131.78 & 199.46 & 0.5262 \\
\bottomrule
\end{tabular}
\end{table}

\subsection{Discussion on the Code Retention Ratio}
\label{app:retention}
The proposed method retains a fixed ratio of teacher codes for student-side quantization, which is set to 50\% for all classification experiments and 12.5\% for all segmentation experiments.
These values were coarsely chosen from a small set of candidate values and then fixed across all settings of each task to avoid extensive scenario-specific tuning.
They should therefore be regarded as practical default operating points rather than individually optimized settings.

Fig.~\ref{fig:selected_code_ablation} shows that the best ratio varies across datasets and transfer directions, indicating that the appropriate amount of retained teacher-side information depends on the distillation scenario.
A smaller ratio may discard useful teacher-side information, whereas a larger ratio may include codes that are less relevant to the task or less compatible with the student modality.
Nevertheless, performance remains relatively stable over a broad range of ratios and generally exceeds the w/o KD baseline.
This indicates that the effectiveness of the proposed method does not depend on a single carefully tuned ratio.

Although the fixed ratio provides consistent improvements, it does not adapt to the varying amount of transferable information across cross-modal scenarios.
Adaptive code retention, which automatically determines the number of retained codes for each scenario, is therefore a promising direction for future work.

\subsection{Class-wise Analysis on NYU-Depth V2}
\label{app:nyu_classwise}
As shown in Table~\ref{tab:seg_result}, the OA of our method in the RGB$\rightarrow$Depth setting is slightly lower than that of the student trained without distillation (0.5218 vs. 0.5269).
To examine whether this decrease reflects negative transfer, we analyze the class-wise recall difference between our method and the student trained without distillation.
Table~\ref{tab:nyu_classwise} reports the five classes with the largest recall increases and decreases.

The analysis shows that the decrease in OA is strongly influenced by class imbalance rather than by a general degradation across classes.
The recall of \texttt{wall} decreases by 0.0274, and since this class occupies 25.6\% of all pixels, this change substantially affects the pixel-weighted OA.
In contrast, several less frequent classes, including \texttt{other\_furniture}, \texttt{mirror}, \texttt{television}, \texttt{desk}, and \texttt{curtain}, show much larger recall improvements of 0.13-0.18, although their contributions to OA are limited by their low pixel frequencies.
Accordingly, CA, which assigns equal importance to each class, improves from 0.2100 to 0.2184 despite the slight decrease in OA, and mIoU also improves from 0.1366 to 0.1456.

These results indicate that the proposed method improves class-balanced performance, and that the OA decrease reflects a class-specific trade-off rather than systematic negative transfer across classes.
The gains on several less frequent classes are also consistent with the benefit of complementary semantic cues provided by the RGB teacher.

\begin{table}[t]
\centering
\small
\caption{Class-wise recall analysis on NYU-Depth V2 in the RGB$\rightarrow$Depth setting.
Freq. denotes the percentage of pixels belonging to each class, and $\Delta$Recall denotes the recall difference between our method and the student trained without distillation.}
\vspace{1mm}
\label{tab:nyu_classwise}
\begin{tabular}{lcc|lcc}
\toprule
\multicolumn{3}{c|}{Top-5 increased classes} & \multicolumn{3}{c}{Top-5 decreased classes}\\
Class & Freq. & $\Delta$Recall & Class & Freq. & $\Delta$Recall\\
\midrule
\texttt{other\_furniture} & 2.7\% & +0.1807 & \texttt{person}    & 0.3\%  & $-$0.1298 \\
\texttt{mirror}           & 1.2\% & +0.1519 & \texttt{counter}   & 1.7\%  & $-$0.0478 \\
\texttt{television}       & 0.7\% & +0.1324 & \texttt{bookshelf} & 2.1\%  & $-$0.0398 \\
\texttt{desk}             & 0.8\% & +0.1315 & \texttt{clothes}   & 0.8\%  & $-$0.0294 \\
\texttt{curtain}          & 1.8\% & +0.1311 & \texttt{wall}      & 25.6\% & $-$0.0274 \\
\bottomrule
\end{tabular}
\end{table}

\subsection{Hard and Soft Code Assignment}
\label{app:soft_assign}
We use a discrete codebook because it provides an explicit and structure-independent concept interface between heterogeneous teacher and student features.
Mapping each feature unit to a finite set of codes allows us to evaluate the contribution of individual concepts, rank them, and retain only the transferable subset.
Such code-level selection is less directly defined when each feature is represented by an unrestricted continuous mixture of anchors.

To examine whether hard assignment to the selected codes is necessary during transfer, we replace the nearest-code assignment in Stage-3 with a soft assignment over the same learned and selected codes.
Specifically, each student unit is represented by a similarity-weighted combination of the selected codes rather than by its single nearest code.
As shown in Table~\ref{tab:soft_assign}, the two variants achieve comparable performance, with hard assignment yielding slightly higher mean accuracy in both transfer directions.
This result shows that the selected codes remain effective under soft assignment, while hard assignment is already sufficient in the current setting.

\subsection{Code Reuse Across Student Units}
\label{app:code_reuse}
The proposed method does not enforce an exclusive one-to-one correspondence between student units and teacher codes.
Multiple student units can be assigned to the same code, allowing a concept to be represented across several temporal or spatial locations.
To quantify this behavior, we measure the proportion of activated codes that are assigned to more than one student unit over the VGG-Sound test set.
We observe that 94.48\% and 97.47\% of the activated codes are reused by multiple student units for A$\rightarrow$V and V$\rightarrow$A, respectively.
This indicates that code usage is predominantly many-units-to-one-code rather than one-unit-to-one-code.

Although code assignment is performed independently for each student unit, the quantized student units retain their temporal or spatial organization and are subsequently processed jointly by the codebook-branch head.
The head can therefore exploit the positions, repetition patterns, and overall arrangement of the assigned codes, which allows concepts distributed across multiple student units to be captured.
However, the current assignment does not explicitly model such distributed concepts before quantization.
Extending the framework to model concepts spanning multiple student units more explicitly is an interesting direction for future work.

\begin{table}[t]
\centering
\small
\caption{Comparison of hard and soft code assignment in Stage-3 on VGG-Sound.
Results are reported as mean $\pm$ standard deviation over five independent runs.}
\vspace{1mm}
\label{tab:soft_assign}
\begin{tabular}{l|cc}
\toprule
Assignment & A $\rightarrow$ V & V $\rightarrow$ A\\
\midrule
Hard & 0.5299$\pm$0.0017 & 0.6229$\pm$0.0059 \\
Soft & 0.5254$\pm$0.0024 & 0.6208$\pm$0.0082 \\
\bottomrule
\end{tabular}
\end{table}

\subsection{Qualitative Analysis of Learned Codes}

\begin{figure*}[t]
\centering

\begin{subfigure}[t]{\linewidth}
    \centering
    \includegraphics[width=\linewidth]{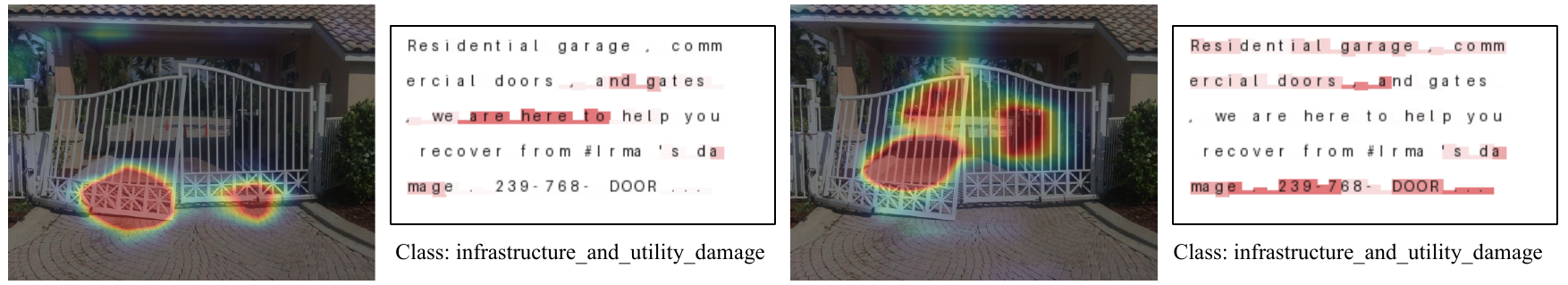}
    \caption{CrisisMMD}
    \label{fig:codebook_qualitative_crisis}
\end{subfigure}

\vspace{2mm}

\begin{subfigure}[t]{\linewidth}
    \centering
    \includegraphics[width=\linewidth]{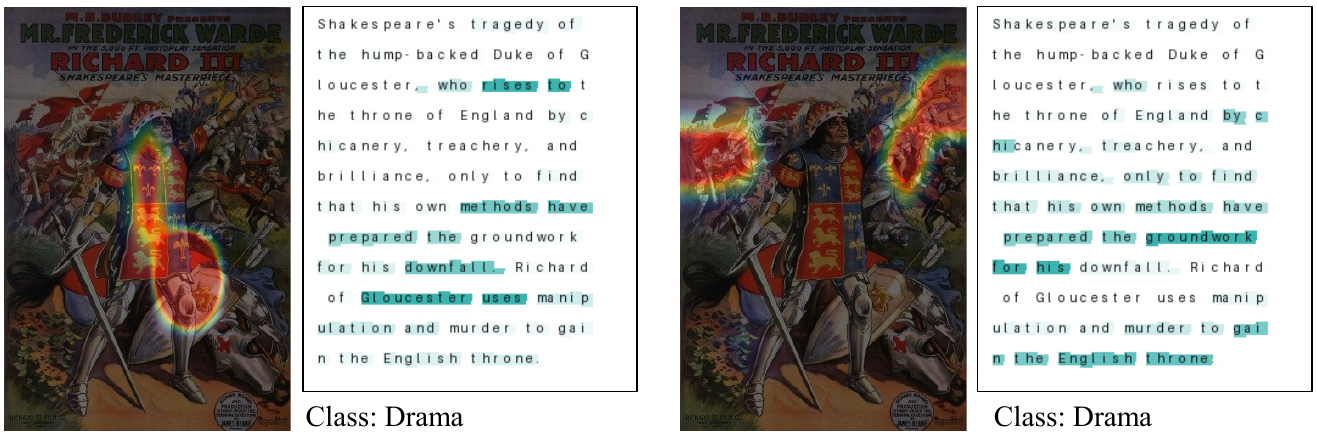}
    \caption{MM-IMDB}
    \label{fig:codebook_qualitative_mmimdb}
\end{subfigure}

\caption{
    Qualitative analysis of learned codes.
    For each selected image-teacher code, the pair shows its associated image region and corresponding high-weight tokens from the distilled text student.
}
\label{fig:codebook_qualitative}
\end{figure*}

Figs.~\ref{fig:codebook_qualitative_crisis} and~\ref{fig:codebook_qualitative_mmimdb} provide qualitative examples of the learned codes on CrisisMMD and MM-IMDB, respectively.
We consider image-teacher distillation settings and visualize selected image-teacher codes together with high-weight tokens from the distilled text student.
For each selected code, we show its associated image regions and the corresponding token weights from the text student.

In the CrisisMMD example, the sample belongs to the \textit{infrastructure\_and\_utility\_damage} class.
The selected image-teacher codes are associated with regions around damaged door or gate structures.
On the text side, the distilled student assigns relatively high weights to related textual contexts, including tokens such as \textit{damage}, \textit{doors}, \textit{gates}, and \textit{DOOR}.
In the MM-IMDB example, the sample belongs to the \textit{Drama} class.
The selected codes attend to visually salient regions in the movie poster, while the text student places relatively high weights on narrative phrases related to dramatic content, such as \textit{tragedy}, \textit{methods}, \textit{prepared the groundwork}, \textit{downfall}, and \textit{English throne}.
These examples suggest that the learned codes can be associated with semantically related visual and textual cues, providing qualitative support for code-based cross-modal transfer.

\newpage
\section{Pseudo Algorithm of The Proposed Framework}
\label{appendix:algorithm}
\begin{algorithm}[t]
\caption{Codebook-Guided Cross-Modal Knowledge Distillation}
\label{alg:proposed}
\begin{algorithmic}[1]
\Require Paired data $\{(\mathbf{x}_i^{(t)},\mathbf{x}_i^{(s)},\mathbf{y}_i)\}_{i=1}^{N}$, pretrained teacher $\{f^{(t)}(\cdot;\theta^{*(t)}),g^{(t)}(\cdot;\phi^{*(t)})\}$, student $\{f^{(s)}(\cdot;\theta^{(s)}),g^{(s)}(\cdot;\phi^{(s)})\}$, codebook size $B$, number of selected codes $K$, code dimension $D$.
\vspace{2mm}
\Ensure Trained student parameter $\theta^{(s)}, \phi^{(s)}$
\vspace{2mm}
\Statex \textbf{Stage-1: Teacher-side codebook learning}
\vspace{1mm}
\State Freeze the pretrained teacher parameters $\theta^{*(t)}, \phi^{*(t)}$.
\State Initialize codebook $\mathbf{C}\in\mathbb{R}^{B\times D}$, projections $\Pi_{\rm in}^{(t)},\Pi_{\rm out}^{(t)}$, and codebook branch head $g_{\rm cp}^{(t)}(\cdot; \phi_{\rm cp}^{(t)})$.
\For{each training iteration}
    \For{$i=1$ to $N$}
        \State Extract teacher features $\mathbf{z}_i^{(t)}=f^{(t)}(\mathbf{x}_i^{(t)};\theta^{*(t)})$.
        \State Quantize teacher features with $\mathbf{C}$ and compute codebook-induced prediction $\mathbf{q}_i^{(t)}$.
        \State Compute student prediction $\mathbf{p}_i^{(s)}$.
    \EndFor
    \State Compute loss according to Eq.~\ref{eq:phase1}.
    \State Update $\mathbf{C}, \Pi_{\rm in}^{(t)}, \Pi_{\rm out}^{(t)}, \phi_{\rm cp}^{(t)},\theta^{(s)}, \phi^{(s)}$.
\EndFor
\vspace{2mm}
\Statex \textbf{Stage-2: Code-wise importance estimation}
\vspace{1mm}
\For{each code $k\in\{1,\ldots,B\}$}
    \State Remove the $k$-th code to construct $\mathbf{C}_{-k}$.
    \State Compute leave-one-code-out teacher prediction $\bar{\mathbf{q}}_i^{(t)}$.
    \State Compute full codebook teacher prediction $\mathbf{q}_i^{(t)}$.
    \State Compute student prediction $\mathbf{p}_i^{(s)}$.
    \State Estimate task relevance $u_k^{\rm rel}$ and student compatibility $u_k^{\rm comp}$ according to Eq.~\ref{eq:score_relevance}.
    \State Compute final importance score $u_k = u_k^{\rm rel} + u_k^{\rm comp}$.
\EndFor
\State Select the top-$K$ codes according to $\{u_k\}_{k=1}^{B}$ and form $\mathbf{C}_{\rm sel}$.
\vspace{2mm}
\Statex \textbf{Stage-3: Selective codebook-based distillation}
\vspace{1mm}
\State Initialize student-side projections $\Pi_{\rm in}^{(s)},\Pi_{\rm out}^{(s)}$ and student-side codebook branch head $g_{\rm cp}^{(s)}(\cdot; \phi_{\rm cp}^{(s)})$.
\State Re-initialize student model parameters $\theta^{(s)}, \phi^{(s)}$.
\State Fix the selected teacher codebook $\mathbf{C}_{\rm sel}$.
\For{each training iteration}
    \For{$i=1$ to $N$}
        \State Extract student features $\mathbf{z}_i^{(s)}=f^{(s)}(\mathbf{x}_i^{(s)};\theta^{(s)})$.
        \State Quantize student features with $\mathbf{C}_{\rm sel}$ and compute student-side codebook-induced prediction $\mathbf{q}_i^{(s)}$.
        \State Compute teacher-side codebook-induced prediction $\mathbf{q}_i^{(t)}$.
        \State Compute student task prediction $\mathbf{p}_i^{(s)}$.
    \EndFor
    \State Compute loss according to Eq.~\ref{eq:phase3}.
    \State Update the student model parameters $\theta^{(s)}, \phi^{(s)}$.
\EndFor

\State \Return Trained student parameter $\theta^{(s)}, \phi^{(s)}, \Pi_{\rm in}^{(s)},\Pi_{\rm out}^{(s)}, \phi_{\rm cp}^{(s)}$.
\end{algorithmic}
\end{algorithm}
Algorithm~\ref{alg:proposed} provides the pseudo-code of the proposed framework.
The procedure consists of three stages, corresponding to the method description in Sec.~\ref{sec:method}: teacher-side codebook learning, code-wise importance estimation, and selective codebook-based distillation. Although the pseudo-code is written at the dataset level for clarity, all optimization steps are implemented with mini-batch training.

\section{Broader Impact}
This work studies a general cross-modal knowledge distillation framework and is not tied to a specific deployment scenario. 
A potential positive impact is that the method may improve efficient model learning by transferring useful knowledge from auxiliary modalities to compact or single-modality students. 
However, as with other distillation methods, the student may inherit biases, spurious correlations, or undesirable behavior from the teacher model and training data. 
Therefore, deployment in sensitive applications should require careful evaluation of fairness, privacy, and robustness under the target use case.

\end{document}